%% file: oshir_focal.tex
\pdfoutput=1
\documentclass{llncs}
\usepackage{amsmath,epsfig,graphicx,amssymb,multicol,psfrag,url,enumerate}
\usepackage{algorithm}
\usepackage{algorithmic}
\usepackage{float}
 \floatstyle{ruled}
\newfloat{algorithm}{tbp}{loa}
\floatname{algorithm}{Algorithm}

\begin{document}
\thispagestyle{empty}
\title{Evolutionary Hessian Learning:\\ Forced Optimal Covariance Adaptive Learning (FOCAL)}
\titlerunning{FOCAL: Forced Optimal Covariance Adaptive Learning}
\author{Ofer M. Shir\inst{1}, Jonathan Roslund\inst{1}, \\ Darrell Whitley\inst{2}, and Herschel Rabitz\inst{1}}
\authorrunning{Shir et al.} 
\institute{Department of Chemistry, Princeton University, USA\\
\medskip
\and
Department of Computer Science, Colorado State University, USA\\
\medskip
\email{oshir@Princeton.EDU}
}
\maketitle
\begin{abstract}
The Covariance Matrix Adaptation Evolution Strategy (CMA-ES) 
has been the most successful Evolution Strategy at exploiting covariance information; it 
uses a form of Principle Component Analysis which, under
certain conditions, is suggested to converge to the correct covariance matrix, formulated as the inverse of the mathematically well-defined Hessian matrix. 
However, in practice, there exist conditions where CMA-ES converges to the global optimum (accomplishing its primary goal) while it does not
learn the true covariance matrix (missing an auxiliary objective), likely due to step-size deficiency. 
These circumstances can involve high-dimensional landscapes ($n\gtrsim 30$) with large condition numbers ($\xi\gtrsim 10^4$). 
This paper introduces a novel technique entitled Forced Optimal Covariance Adaptive Learning (FOCAL), with the explicit goal of determining the Hessian at the global basin of attraction.
It begins by introducing theoretical foundations to the inverse relationship between the learned covariance and the Hessian matrices.
FOCAL is then introduced and demonstrated to retrieve the Hessian matrix with high fidelity on both model landscapes and experimental Quantum Control systems, 
which are observed to possess a non-separable, non-quadratic search landscape. 
The recovered Hessian forms are corroborated by physical knowledge of the systems.
This study constitutes an example for \emph{natural computing} successfully serving other branches of natural sciences, 
and introducing at the same time a powerful generic method for any high-dimensional continuous search seeking landscape information.
\end{abstract}


\section{Introduction}
\begin{figure}
\centering \includegraphics[scale=0.6]{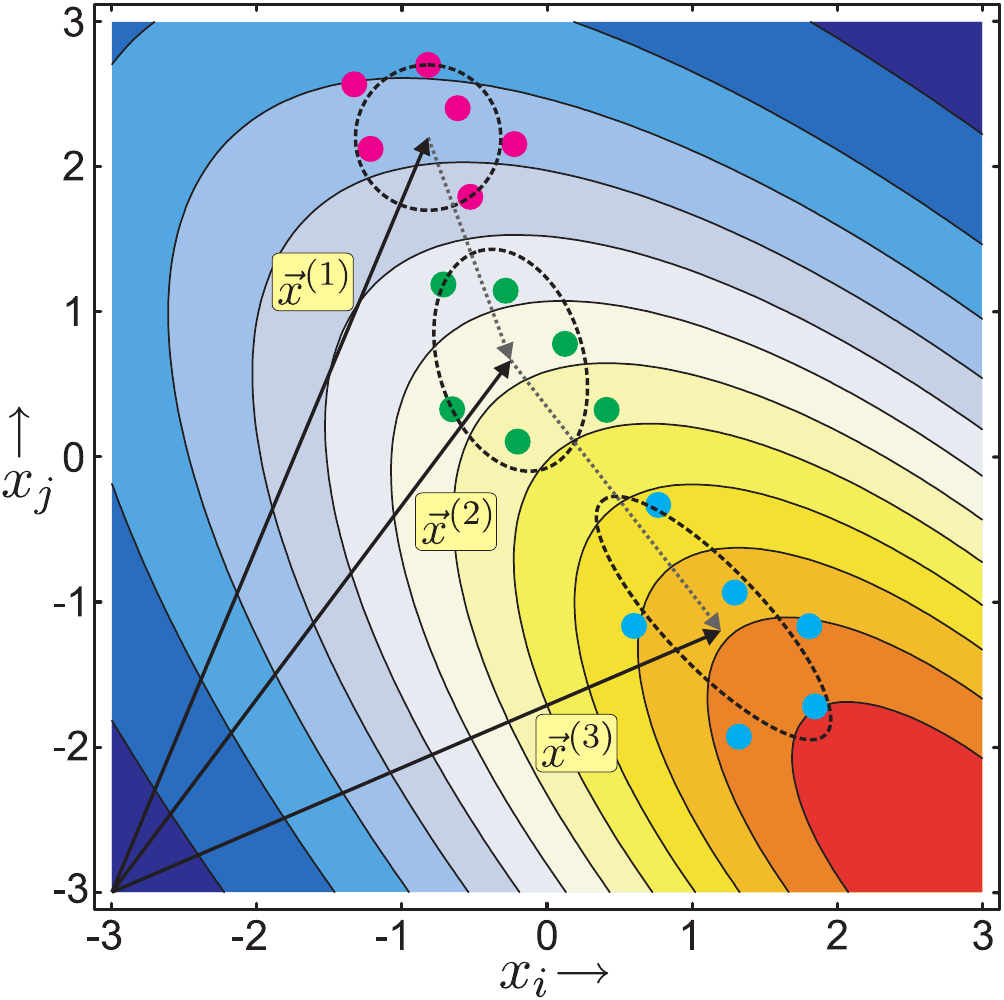}
\caption{CMA-ES optimization on a 2-dimensional model landscape.
The various ellipsoids define \emph{equiprobability} sampling contours at
different locations on the search landscape. Principal Component
Analysis of directions of the highest yield improvements
allows for continual refinement and coordinate rotation of these
distributions in order to generate an optimal sampling distribution. \label{fig:cmacartoon}}
\end{figure}

Automated learning of the Hessian matrix about the global optimum in the presence of noise and without any derivative measurement constitutes 
a challenging task. 
Knowledge of second-order information at the optimum is very desirable as it may offer a measure of \emph{system robustness}, 
e.g., to noise, allow for dimensionality reduction, and assist in landscape characterization.
Within the Computational Intelligence community, 
Hessian learning is occasionally associated with the operation of Evolution Strategies (ES) \cite{Beyer-Schwefel}, 
powerful bio-inspired search heuristics, 
which excel in high-dimensional search of continuous landscapes. 
The advent of modern ES, also known as Derandomized Evolution Strategies (DES) \cite{Hansen01completely},
facilitates efficient global optimization with small population sizes, subject to real-time statistical learning of the successful consecutive steps.
Following a series of successful DES variants, formulated throughout the early 1990's \cite{HansenDR1,HansenDR2,HansenDR3}, 
the Covariance Matrix Adaptation Evolution Strategy (CMA-ES) emerged \cite{HansenDR4,hansencmamultimodal} 
as a state-of-the-art Evolution Strategy. It has become a successful global optimizer, 
outperforming other stochastic search heuristics in various reported competitions (see, e.g., \cite{HansenCEC2005b,HansenGECCO09}), 
being increasingly employed in a long list of real-world applications \cite{CMAapplications}, and having extensions to multi-objective 
Pareto optimization \cite{CMA-MO}, uncertainty handling \cite{hansen2009tec}, niching \cite{Shir-NACO08,Shir-SA_ECJ}, to mention a few.
At the same time, this popular algorithm was also simplified in a competing strategy (the CMSA; see \cite{BeyerCMA}), 
and was even further improved for certain cases of global optimization \cite{Akimoto}.
The CMA-ES targets an efficient on-the-fly learning of the optimal mutation distribution by applying 
\emph{Principal Component Analysis} (PCA) to the successful search-variations within a population of candidate solutions. 
Toward this end, it constructs a covariance matrix that uniquely dictates the operation of the mutation mechanism
(see Fig.\ \ref{fig:cmacartoon} for an illustration).
Most ES, and particularly the CMA-ES, have long aimed to learn a covariance matrix associated with the Hessian matrix. 
Although this behavior has never been rigorously addressed, it has been implicitly expressed 
in conventional ES literature (see, e.g., \cite{CMAES-Tutorial}). 
A crucial question is whether the constructed covariance matrix is always related to the mathematically well-defined Hessian matrix, or rather
simply reflects the point dispersion of a stochastic search path guided by selection pressure.
Another issue is whether there are scenarios where the CMA-ES accomplishes convergence to the global optimum, 
while it does not learn the true covariance matrix during this process.
The goal of the current work is thus three-fold: 
\begin{enumerate}
 \item Establish rigorous principles for the relationship between the inverse Hessian matrix and the covariance matrix 
that is learned by an Evolution Strategy.
\item Illustrate and explain scenarios where the CMA-ES does not learn a covariance matrix reflective of the inverse Hessian matrix at the 
global optimum.
\item Present a new method, relying on existing DES mechanisms but rather possessing a shifted focus toward landscape learning, with 
the explicit goal of Hessian determination at the global basin of attraction. This novel technique, entitled Forced Optimal Covariance 
Adaptive Learning (FOCAL), primarily targets practical \emph{experimental optimization} scenarios. 
\end{enumerate}

The specific application that motivated the development of FOCAL is the Quantum Control (QC) field,
which aims at coherently altering quantum dynamics phenomena typically by means of optimally shaped ultrafast laser fields
\cite{Hersch93,Hersch00,Gerber07}. Following the formulation of Quantum Control Theory (QCT) \cite{Hersch88}, its laboratory realization was achieved
by means of closed-loop experiments \cite{Hersch92} leading to a growing list of 
applications in a broad range of research topics (see \cite{Gerber07} and references therein). 
In recent years the interest in the QC field grew also within the Computational Intelligence community, likely due to the
popularity of Evolutionary Algorithms as the routinely employed optimization heuristics in closed QC learning loops 
(see, e.g., \cite{baumert97,zemo01,pearson01,LabES}), in parallel to the growing understanding of QC
search landscapes \cite{Hersch06,Raj07}.
Furthermore, in the context of the current study, various DES routines also started to be deployed in the optimization of QC systems
\cite{Christian-Elsevier,SHIR_CEC06,Shir-JPhysB}, along with the successful application of CMA-ES in experimental QC single-objective \cite{QCE_GECCO08,LarsJennifer,BartelsCMA} and multi-objective \cite{ShirGecco09} systems.


The remainder of the manuscript is organized as follows.
Section \ref{sec:DES} will introduce the fundamental concepts behind the process of covariance learning and Hessian determination. 
It will review the ES machinery, especially in light of the mutation operator and covariance matrix learning, 
and outline the theoretical principles for the inverse relation between the covariance and the Hessian matrices.
Section \ref{sec:focal} will discuss the limitations of Evolution Strategies, and in particular of the CMA-ES, to 
accomplish successful learning of the inverse Hessian matrix, and will illustrate this behavior under certain conditions. 
We shall then present our proposed scheme, the FOCAL algorithm, and discuss its mechanism in detail.
A preliminary proof of concept will be then demonstrated on a noisy model landscape.
This will be followed by the description of the experimental systems under study in Section \ref{sec:systems} and the practical observation
in Section \ref{sec:observation}. We will discuss our results and summarize this work in Section \ref{sec:discussion}.

\section{From Covariance Learning to Hessian Determination}
\label{sec:DES}
This section will provide some background on the ES machinery, and especially on the mutation operator and its
statistically learned covariance matrix. We will then examine the foundation of the inverse relationship
between the covariance matrix and the Hessian matrix at the top of the landscape.
\subsection{Covariance Utilization in Evolution Strategies}  
\label{sec:ESMutation}
\begin{figure}
 \centering
\includegraphics[scale=0.7]{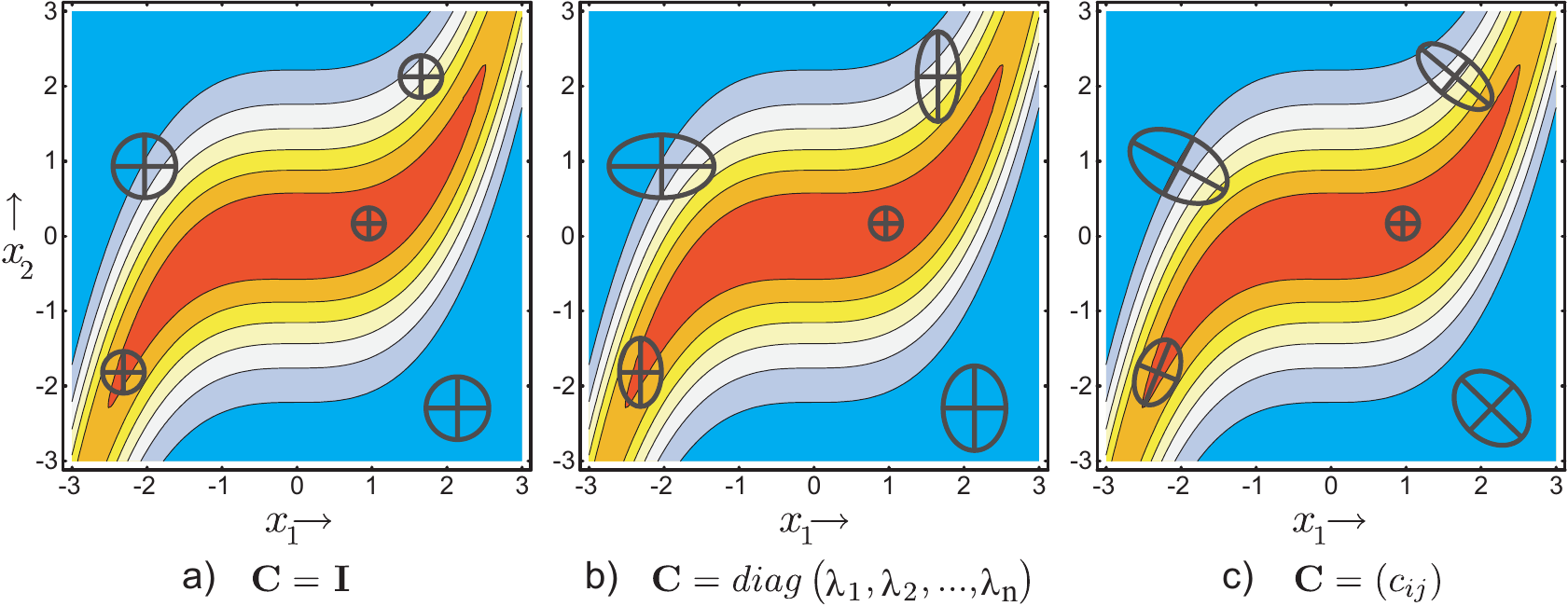}
\caption{Equidensity mutation sampling ellipsoids corresponding to three covariance matrix forms in ES, depicted on a 2-dimensional model landscape.
Panel (a) displays a constant identity matrix. Panel (b) uses $\mathcal{O}(n)$ independent parameters to generate 
on-axis ellipsoids.  Panel (c) utilizes $\mathcal{O}(n^2)$ parameters to create an arbitrary covariance matrix. \label{fig:ESMutSchemes}}
\end{figure}

ES uses stochastic continuous variations for the mutation operation
which are based upon the multivariate normal distribution. The latter is well defined by a covariance matrix.
Furthermore, in non-elitist DES strategies, every generation concludes with a single search point, referred to as the parent,
either by means of a single selection in a $\left(1,\lambda\right)$-strategy, or by means of recombination in
a $\left(\mu,\lambda\right)$-strategy.
Let the search seek the global maximum in a landscape of dimension $n$.
Given the parent in generation $g$, $\vec{x}^{(g)}$, DES generate offspring 
$\vec{x}_k^{(g+1)}$, $k=1\ldots\lambda$, according to the following characteristic equation:
\begin{equation}
 \vec{x}_k^{(g+1)} = \vec{x}^{(g)} + \sigma^{(g)} \cdot \mathcal{N}_k\left(\vec{0},\mathbf{C}^{(g)} \right)~~~~~k=1\ldots\lambda,
\end{equation}
where $\sigma^{(g)}$ is the global step-size, and $\mathbf{C}^{(g)}$ is a covariance matrix defining the mutation distribution.
There are three canonical correlation mutation schemes, corresponding to three distinct forms of the covariance matrix
(for a broad overview we refer the reader to \cite{Baeck-book};
note that in our notation, unlike the Standard ES, the global step-size and the covariance matrix are considered as separate entities): 
\begin{enumerate}[(A)]
\item A constant identity matrix, i.e., adaptation exclusively relies on a step-size mechanism to update the global step-size:
\begin{equation} \label{eq:Ca} \mathbf{C}_a=\mathbf{I} = \textrm{const}\end{equation}
\item A diagonalized covariance matrix, i.e., a vector of $n$ independent strategy parameters,
\emph{presumably} comprising the variances of the decision variables, typically referred to as the \emph{individual step-sizes}:
\begin{equation} \label{eq:Cb} \mathbf{C}_b = \textrm{diag} \left(\lambda_1,\lambda_2,\ldots,\lambda_n\right)=
\textrm{diag} \left(\textrm{VAR}(x_1),\textrm{VAR}(x_2),\ldots,\textrm{VAR}(x_n)\right)  
\end{equation}
\item A general non-singular covariance matrix with arbitrary $\left(n\cdot(n+1)\right)/2$ independent strategy parameters:
\begin{equation} \label{eq:Cc} \mathbf{C}_c= \left(c_{ij}\right) = \mathbf{R} \mathbf{\Lambda} \mathbf{R}^{T},\end{equation}
introducing an eigendecomposition by means of an orthonormal coordinate rotation matrix $\mathbf{R}$ and the eigenvalue matrix
$\mathbf{\Lambda}$.
\end{enumerate}
A geometrical interpretation of the three schemes is given in Fig.\ \ref{fig:ESMutSchemes}.
In the context of this study, only case (C) may offer an inversion to the fully-formed Hessian matrix, 
whereas case (B) may possibly constitute a compressed inverse form of the Hessian eigenvalues.
Upon realization of these three schemes, this study will consider the default CMA-ES \cite{hansencmamultimodal} for case (C) (denoted by
def-CMA-ES), the sep-CMA-ES \cite{HansenDR2PPSN08} for case (B), and the CMA with a constant unit matrix for case (A); 
The latter is a zero-order DES which generates isotropic mutations and employs the Cumulative Step-Size Adaptation (CSA) mechanism 
\cite{Ostermeier94} as its adaptation mechanism, 
and will be referred here as the iso-CMA-ES. Moreover, we shall consider the $\left(\mu_W,\lambda\right)$ strategy, 
i.e., a non-elitist strategy generating $\lambda$ search-points, amongst which it recombines the $\mu$ with the highest ranking in a weighted averaging.

\subsection{Covariance versus Hessian: Principles of Inversion}
Since the development of Evolution Strategies, it has been implicitly assumed that the evolving covariance matrix approximates the inverse Hessian 
of the search landscape. This was primarily supported by the rationale that locating the global basin of attraction by an ES can be accommodated
using mutation steps that fit the actual landscape, or, equivalently, that the optimal covariance distribution 
can offer mutation steps whose \emph{equidensity probability contours} match the \emph{level sets} of the landscape \cite{Rudolph92}.
Another rationale behind this was the following: 
the reduction of a general problem to a \emph{spherical} problem may be achieved by sampling search-points based upon a covariance matrix 
that is the inverse Hessian (note that this is equivalent to replacing the Euclidean metric by the Mahalanobis metric \cite{Shir-SA_ECJ}).  
Since ES optimally operate on the quadratic sphere model, any successful ES run suggests that the covariance learning was accomplished.
However, at the same time, there has been no rigorous analysis that explicitly demonstrates 
that ES machinery indeed learns the inverse of the Hessian.
A study by Rudolph \cite{Rudolph92} proved the potential of ES to facilitate such learning
and derived practical bounds on the population size toward the end of successful learning. The latter work also questioned the effectiveness of ES learning
upon discarding past information and claimed that ES learning would benefit from introducing memory to the individuals. In retrospect, this
has indeed been achieved throughout the development of DES, which provides the basis to introduce FOCAL in the current study.

We now analyze the inverse relation between the covariance and the Hessian matrices at the top of the landscape.
This section will thus focus on the covariance matrix upon reaching the proximity of the global attraction basin, when subject to 
a self-adaptive Evolution Strategy.

A covariance matrix amongst sets of variates is defined as
\begin{equation*}
\mathcal{C}_{ij}=\left< x_i x_j \right> - \left< x_i\right> \left<x_j \right>.
\end{equation*}
By construction, the origin is set at the parent search-point, which is located at the optimum.
The covariance elements are then reduced to the following \emph{expectation values}:  
\begin{equation}
\label{eq:Cov0}
\mathcal{C}_{ij}=\left< x_i x_j \right> = \int_{-\infty}^{+\infty} x_i x_j \mathcal{P}\left( \vec{x}\right) d\vec{x},
\end{equation}
where $\mathcal{P}\left( \vec{x}\right)$ is the probability distribution function (PDF), 
and the variables $x_i$ correspond to point dispersion about the optimum.
Revealing the nature of the PDF is necessary for the interpretation of the covariance matrix.
Here, the sampling of the variate will be dictated by the selection mechanism of the Evolution Strategy, and upon   
considering a deterministic non-elitist $\left(\mu,\lambda\right)$-strategy, the PDF depends only upon the objective function value. 
Since we assume that neither penalties nor cost functions are employed, the PDF becomes the following functional of the yield: 
\begin{equation}
 \mathcal{P}\left( \vec{x}\right) \equiv \mathcal{P}\left[ J\left( \vec{x}\right)\right].
\end{equation}
$J\left( \vec{x}\right)$ refers here to the distance in the \emph{objective space} of the trial solution, 
$\vec{x}$, to the global optimum location $\vec{x}^{*}$,
\begin{equation*}
 J\left( \vec{x}\right) = f_{\max}-f\left( \vec{x}\right) = f\left( \vec{x}^{*}\right) - f\left( \vec{x}\right)
\end{equation*}
i.e., it does not refer to the absolute yield value, but rather to possible yield deviations from the optimum. 

Importantly, the selection mechanism is blind to the location of the candidate solutions
in the search space, and its solitary criterion is the ranked yield values.
Consequently, the PDF is a function of the objective value alone,
\begin{equation*}
\mathcal{P} \equiv \mathcal{P} \left( J \right).
\end{equation*}
{\bf The crucial claim is that $\mathcal{P} \left( J \right)$ follows the exponential distribution:}
\begin{equation}
 \mathcal{P} \left( J \right) \sim \gamma \exp\left(-\gamma J \right).
\end{equation}
It is important to note that while the line of reasoning leading to this distribution is general, the exponential distribution itself is strictly valid only in the immediate vicinity of the optimum.
Elsewhere, the mutation-selection probability distribution must be weighted by the probability of generating a given mutation; this supplemental probability distribution is monotonically decreasing with the yield $J$ 
only at the optimum, which prevents it from significantly distorting the exponential selection distribution. 
In what follows, we show the correctness of the claim from two different perspectives.
\subsubsection*{The Memoryless Property}
Given the deterministic ranking of the $\lambda$ candidate solutions per generation, $\left\{ \vec{x}_{k:\lambda}\right\} _{k=1}^{\lambda}$,
based upon their ascending ranked distance to the global optimum $\left\{ J_{k:\lambda}\right\} _{k=1}^{\lambda}$,
we examine the selection probability $\mathcal{P} \left( J_{i:\lambda} \right)$ corresponding to the trial solution $\vec{x}_{i:\lambda}$.
Upon determination of the best individual in the current generation, $\vec{x}_{1:\lambda}$, the selection probability of the $i^{th}$
individual depends only upon its relative distance in the objective space to $J_{1:\lambda}$, denoted as $\Delta J_{i,1}=J_{i:\lambda}-J_{1:\lambda}$:
\begin{equation*}
 \mathcal{P}\left( J_{i:\lambda} \left|  J_{1:\lambda}\right. \right) = \mathcal{P}\left( \Delta J_{i,1} \right).
\end{equation*}
Consider Bayes' theorem,
\begin{equation*}
 \mathcal{P}\left( J_{i:\lambda} \left|  J_{1:\lambda}\right. \right) = \mathcal{P}\left( J_{1:\lambda} \left|  J_{i:\lambda}\right. \right) \cdot
\frac{\mathcal{P}\left( J_{i:\lambda}\right)}{\mathcal{P}\left( J_{1:\lambda}\right)},
\end{equation*}
and note that $\mathcal{P}\left( J_{1:\lambda} \left|  J_{i:\lambda}\right. \right)=1$, which allows us to conclude 
\begin{equation}
\label{eq:memoryless}
 \mathcal{P}\left( J_{i:\lambda} \right) = \mathcal{P}\left( \Delta J_{i,1} \right)\cdot\mathcal{P}\left( J_{1:\lambda} \right).
\end{equation}
Note that this functional equation is only satisfied with the exponential distribution.
In other words, the selection probability for a given individual has no memory of its absolute distance to the global optimum,
but rather depends upon its relative distance to other candidate solutions of the same generation, or equivalently, we may state that the variable
$J_{i:\lambda}$ is \emph{memoryless} with respect to the global optimum $J^{*}=0$. 
Since {\bf the exponential distribution is the only continuous memoryless random distribution}, 
$\mathcal{P} \left( J \right)$ must necessarily follow it. 

\subsubsection*{The Defining Equation}
Upon increasing the value of $J$, the corresponding rank of the trial solution decreases. 
This directly translates into the requirement that the probability distribution function should be monotonically decreasing when 
increasing $J$ values, i.e., $\frac{d\mathcal{P}}{dJ} < 0$. A simple way to satisfy this requirement is by considering
its derivative by means of a positive weighting function $w\left(J\right)$:
\begin{equation}
 \frac{d\mathcal{P}}{dJ} = -\nu w\left(J\right),
\end{equation}
with the following \emph{boundary conditions},
\begin{equation}
\begin{array}{l}
\medskip
\displaystyle w\left(J=0\right)=1\\
\displaystyle w\left(J\rightarrow \infty \right) \rightarrow 0, 
\end{array}
\end{equation}
and where the scalar $\nu$ accounts for the appropriate scaling.
Note that a Gaussian profile is already ruled out at this point, since its first derivative vanishes at the origin.
In addition, the weighting function $w\left(J\right)$, which may be interpreted as the \emph{mutation sensitivity},
is monotonically decreasing ($\frac{dw}{dJ} < 0$) according to the following qualitative rationale:
larger yield decrease is more likely to occur upon perturbing higher ranked solutions, i.e. low $J$ values, 
when compared to perturbation of lower ranked solutions. 
Following the same arguments upon which we constructed $w\left(J\right)$, we may generate 
an equivalent weighting function with the appropriate properties: 
\begin{equation*}
 \frac{dw}{dJ} = -\eta \tilde{w}\left(J\right).
\end{equation*}
A simple way to satisfy these demands is by setting:
\begin{equation}
 w\left(J\right) = \tilde{w}\left(J\right) ~ \Longrightarrow ~ w\left(J\right) \sim \exp \left( -\eta J \right).
\end{equation}
These recursive relations lead us to conclude that the parent function, $\mathcal{P} \left( J \right)$ is the exponential function,
which is known to be the only mathematical function constituting its own derivative:
\begin{equation}
\mathcal{P} \left( J \right) \sim \exp\left(-\nu J \right).
\end{equation}

\subsection*{Obtaining the Hessian}
Given the conclusions of the previous section, we may rewrite Eq.\ \ref{eq:Cov0} as follows:
\begin{equation}
 \mathcal{C}_{ij} = \gamma \int_{-\infty}^{+\infty} x_i x_j \exp\left[-\gamma J\left( \vec{x}\right)\right] d\vec{x}.
\end{equation}
We focus here on the attractive basin of the global optimum,
where the PDF allows continuous sampling while maintaining the yield values at the top of the landscape. 
Hence, $J\left( \vec{x}\right)$ may be {\bf Taylor-expanded} about the optimum,
\begin{equation}
 J\left( \vec{x}\right) \simeq \frac{1}{2} \vec{x}^T \cdot \mathcal{H} \cdot \vec{x},
\end{equation}
which in combination with the normalized form of the exponential distribution yield:
\begin{equation}
 \mathcal{P}\left[ J\left( \vec{x}\right)\right] \simeq \left(\frac{\gamma}{2\pi} \right)^{n/2} \left| \mathcal{H} \right|^{1/2}
\exp \left( -\frac{1}{2}\gamma \vec{x}^T \cdot \mathcal{H} \cdot \vec{x} \right)
\end{equation}
where $\left| \mathcal{H} \right|$ is the Hessian's determinant.
Upon substitution to Eq.\ \ref{eq:Cov0}, the covariance element reads
\begin{equation}
 \mathcal{C}_{ij}=\left(\frac{\gamma}{2\pi} \right)^{n/2} \left| \mathcal{H} \right|^{1/2} \int_{-\infty}^{+\infty} x_i x_j
\exp \left( -\frac{1}{2}\gamma \vec{x}^T \cdot \mathcal{H} \cdot \vec{x} \right) d\vec{x}
\end{equation}
Finally, the integration can now be completed to obtain the desired covariance element:
\begin{equation}
 \mathcal{C}_{ij} = \left(\mathcal{H}^{-1}\right)_{ij}
\end{equation}

\section{FOCAL: Forced Optimal Covariance Adaptive Learning}
\label{sec:focal}
The statistical machinery of the CMA-ES has the necessary tools to obtain the
covariance matrix of the decision variables in the presence of noise without any derivative
estimation, and upon inversion, to attain the Hessian matrix at the global optimum. 
If the learning of the covariance matrix is successful upon convergence to the optimum, 
the evolution strategy may continue to generate mutations without affecting the target control yield (fitness),
since the extended sampling results in a large dispersion of points for insensitive
search directions (i.e., small magnitude Hessian eigenvalues). At the same time,
mutations that obtain significant yield deterioration are rejected,
leading to a small point dispersion for highly sensitive search
directions (i.e., large magnitude Hessian eigenvalues). 

The default CMA-ES does not necessarily operate in the manner above, and its covariance matrix is not always able to evolve to a 
form that properly reflects the structure of the underlying search landscape. The rationale for this deviation from its prescribed behavior is two-fold. 
First, at the practical level, the default covariance matrix \emph{learning rate} is proportional to the reciprocal of the squared search-space dimensionality 
and therefore \emph{precludes meaningful learning} in a timely manner for large search space dimensions, e.g., $n \gtrsim 30$. 
More importantly, at a deeper conceptual level, since the covariance matrix adaptation is carried out simultaneously with the
Cumulative Step-Size Adaptation (CSA) scheme \cite{Ostermeier94}, upon approaching a basin of attraction the global step-size shrinks to zero
and the exploration of the landscape is practically halted. 
There exist certain situations (e.g., landscapes with large condition numbers, $\xi \gtrsim 10^4$), 
in which this CSA behavior constitutes an obstacle to the CMA-ES learning the correct covariance matrix.
As demonstrated later in Figures \ref{fig:ellipse} and \ref{fig:SHG_spectrum},
the default CMA-ES does not perform well in identifying a high-dimensional Hessian ($n=80$)
and does not learn the correct eigenspectrum.
These observations correspond to simulations and experiments that were conducted within realistic environments of noisy decision (input)
variables and a practical budget of function evaluations ($\sim 10^4$) and will be described in greater detail in what follows.
 
The main idea behind the newly proposed FOCAL technique is to force
landscape exploration for the sake of statistical analysis, by
preventing this stagnation via the global step-size update scheme.
FOCAL is thus introduced as a modification to the CMA-ES for the sake of landscape learning, 
which can also be utilized in other DES procedures.
It is important to note, as was previously suggested, that FOCAL's success is contingent on the search-engine's
capability to reach the proximity of the global optimum. Failure to meet this criterion would naturally render
FOCAL-based Hessian learning invalid.

In this section we shall consider the CMA-ES as the search-kernel of the FOCAL method, representing
case (C) in the mutation schemes outlined in Section \ref{sec:ESMutation}. Since FOCAL is a generic method, alternative search-kernels may be employed. 
In this work we shall consider two additional kernels. The reduction to
case (B), when a compressed landscape picture is sought in the form of a diagonalized matrix, is straightforward.
Furthermore, the consideration of case (A) (zero-order DES with isotropic mutations) within the FOCAL context 
will constitute a reference routine, i.e., demonstrating the necessity for on-axis or arbitrarily rotated mutations. 
The success or failure to fulfill FOCAL's goals on a certain problem does not 
necessarily reflect the performance of the employed search-engine for the canonical global optimization of the same problem. 
It is possible that an equivalent search with the same kernel but with an alternative step-size mechanism 
would obtain a different outcome, e.g., locating the global optimum.
This differentiation stems from the primary objective of FOCAL, that is, precise learning of the optimal sampling
distribution, rather than yield maximization as in canonical global optimization. 

\subsection{FOCAL: Mechanism Description}
FOCAL introduces two main modifications to the default CMA-ES algorithm:
\begin{itemize}
 \item Adjusting the value of the covariance matrix learning rate, $c_{cov}$
 \item Reformulating the global step-size update scheme  
\end{itemize}
In what follows, we will elaborate on each of those two components.
\subsubsection{The Learning Rate $c_{cov}$}
The default learning rate of the covariance matrix within the CMA-ES is proportional to the reciprocal of the squared
search-space dimensionality, i.e., $c_{cov} \propto 1/n^2$ \cite{hansencmamultimodal}. 
It therefore precludes meaningful learning, especially when the search space dimensionality is large, e.g., $n \gtrsim 30$.  
With this in mind, the learning rate is adjusted in FOCAL to allow contribution of at least several percent (order of $1\% \sim 10\%$,
i.e., $c_{cov} = 0.01 \sim 0.10$) to the covariance update from the statistical analysis of successful mutations.
Also, when considering the auxiliary factor $H_{\sigma}$ in certain implementations \cite{hansencmamultimodal} it should be permanently 
fixed as $H_{\sigma}=1$.
Setting $c_{cov}$ becomes a user task, which should take into account the trade-off
between convergence speed and convergence reliability; lower covariance contributions should allow for more robust convergence at the expense
of additional function evaluations, and \emph{vice versa}. \emph{A priori} knowledge about the problem, such as the estimated rank of the target Hessian,
may assist in this task. See the discussion about parameter settings in Section \ref{sec:parameters}.

\subsubsection{The Practical Step}
Upon updating the covariance matrix, its eigendecomposition yields a
diagonal matrix $\mathbf{\Lambda}^{(g)}$, which comprises the variances of the rotated control variables:
\begin{equation}\mathbf{\Lambda}^{(g)} = \textrm{diag} \left(\lambda_1^{(g)},\lambda_2^{(g)},\ldots,\lambda_n^{(g)}\right),\end{equation}
with the eigenvalues $\left\{ \lambda_\jmath^{(g)} \right\}_{\jmath=1}^n$ sorted, i.e., $\lambda_1^{(g)}=\lambda_{\max}^{(g)}$,
$\lambda_n^{(g)}=\lambda_{\min}^{(g)}$.

Let us consider the following \emph{difference vector}:
\begin{equation}
\begin{array}{c}
 \medskip
\displaystyle \Delta\vec{x} ^{(g)} = \sigma^{(g)} \cdot \mathbf{R}^{(g)} \cdot \left(\mathbf{\Lambda}^{(g)}\right)^{1/2} \cdot \left\langle\vec{z}\right\rangle_W \\ 
\displaystyle  \left\langle\vec{z}\right\rangle_W = \sum_{\imath=1}^{\mu}w_{\imath}\cdot \vec{z}_{\imath:\lambda},~~~~~\vec{z}_{\imath} \sim \mathcal{N}\left(\vec{0},\mathbf{I}\right)

\end{array}
\end{equation}
where $\sigma^{(g)}$ is the global step-size, and $\mathbf{R}^{(g)}$ is the orthonormal coordinate rotation matrix obtained by the routinely employed eigendecomposition of the covariance matrix. 
This vector constitutes the translation between the parent point to its offspring, and deriving its expected behavior will assist in explaining the 
rationale behind the FOCAL mechanism.
We therefore choose to examine the Root-Mean-Square (RMS) statistics of this difference-vector, 
\begin{equation}
\begin{array}{l}
 \medskip
\displaystyle \frac{\left\langle \left\| \Delta \vec{x} ^{(g)}\right\|^2\right\rangle} {\left(\sigma^{(g)}\right)^2} = \frac{\left\langle \left(\Delta \vec{x} ^{(g)}\right)^T \cdot \Delta \vec{x}^{(g)}\right\rangle} {\left(\sigma^{(g)}\right)^2} \\
 \medskip
\displaystyle = \left\langle \left\langle\vec{z}\right\rangle^T_W \cdot \left(\mathbf{\Lambda}^{(g)}\right)^{1/2} \cdot \left(\mathbf{R}^{(g)}\right)^{T} \cdot \mathbf{R}^{(g)} \cdot \left(\mathbf{\Lambda}^{(g)}\right)^{1/2} \cdot \left\langle\vec{z}\right\rangle_W \right\rangle \\ \medskip
\displaystyle = \left\langle \left\langle\vec{z}\right\rangle^T_W \cdot \mathbf{\Lambda}^{(g)} \cdot \left\langle\vec{z}\right\rangle_W \right\rangle = 
\left\langle \sum_{\jmath}^n \lambda^{(g)}_{\jmath} \cdot \left(\left\langle z \right\rangle_W^{(\jmath)} \right)^2 \right\rangle =
\frac{1}{\mu_{\textrm{eff}}}\cdot\sum_{\jmath}^n \lambda^{(g)}_{\jmath} \cdot \left\langle \mathcal{N}_{\jmath}\left(0,1\right)^2 \right\rangle \\
\displaystyle  = \frac{1}{\mu_{\textrm{eff}}}\cdot\sum_{\jmath}^n \lambda^{(g)}_{\jmath} = \frac{1}{\mu_{\textrm{eff}}}\cdot\textrm{tr}\left(\mathbf{\Lambda}^{(g)} \right) = 
\frac{1}{\mu_{\textrm{eff}}}\cdot\textrm{tr}\left(\mathbf{C}^{(g)} \right)
\end{array}
\end{equation}
where $\mu_{\textrm{eff}} = 1/\sum_{\imath=1}^{\mu}w_{\imath}^2$ \cite{hansencmamultimodal}, and the recombined cross-terms 
vanish\footnote{Since the sampled random variables are mutually independent, we have 
$$\left\langle \left(\sum_{k=1}^{\mu}w_{k}z_{k}^{(\jmath)}\right)^2 \right\rangle = 
\left\langle \sum_{k=1}^{\mu}\sum_{l=1}^{\mu}w_{k}w_{l}z_{k}^{(\jmath)}z_{l}^{(\jmath)} \right\rangle =
\left\langle \sum_{k=1}^{\mu}w_{k}^2\left(z_{k}^{(\jmath)}\right)^2 \right\rangle = \sum_{k=1}^{\mu}w_{k}^2 = \frac{1}{\mu_{\textrm{eff}}}$$}. 
Also, note that the trace is \emph{preserved} through the eigendecomposition.

Since $\left\| \Delta \vec{x} ^{(g)}\right\| \simeq \left\|\Delta\vec{x}^{(g)}\right\|_{RMS}$, we then {\bf define} the practical step-size as the RMS:
\begin{equation}
\label{eq:pstepm}
\displaystyle \delta_p^{(g)} \equiv \left\|\Delta\vec{x}^{(g)}\right\|_{RMS} = 
\sqrt{\left\langle \left\| \Delta \vec{x} ^{(g)}\right\|^2\right\rangle}
 = \frac{\sigma^{(g)}}{\sqrt{\mu_{\textrm{eff}}}}\cdot \sqrt{\textrm{tr}\left(\mathbf{C}^{(g)}\right)}
\end{equation}
This scalar thus reflects the interplay between the global step-size and the effective sampling coverage, 
which is represented by the eigenvalue spectrum of the covariance matrix. In what follows we shall consider $\mu_{\textrm{eff}}=1$.

Bounding the practical step-size is of particular interest here. 
Subject to any step-size mechanism, the upper and lower bounds of the practical step-size may be inferred from Eq.\ \ref{eq:pstepm} 
upon consideration of extreme scenarios of isotropic covariance matrices with radii of either $\lambda_{\min}$ or $\lambda_{\max}$: 
\begin{equation}
\label{eq:bounds-pstepm}
\displaystyle \sigma^{(g)} \cdot \sqrt{n\cdot\lambda_{\min}^{(g)}} \lesssim \delta_p^{(g)} \lesssim 
\sigma^{(g)} \cdot \sqrt{n\cdot\lambda_{\max}^{(g)}}.
\end{equation}
At the top of the landscape, stable convergence is possible with a well adapted global step-size, 
which is likely to shrink to zero by means of the step-size mechanism -- e.g., the CSA mechanism -- 
regardless of the learned sampling distribution.
The crucial point is that the practical step-size $\delta_p^{(g)}$ will vanish even with a poor sampling 
distribution around the optimum, as it is primarily governed by the global step-size: 
\begin{equation}
  \displaystyle \sigma^{(g)}\xrightarrow[J\rightarrow 0]{} 0 ~~\Longrightarrow ~~
\delta_p^{(g)}\xrightarrow[J\rightarrow 0]{} 0,
\end{equation}
where $J$ denotes the distance to the global optimum in the objective space, as before.
We stress that this does not constitute any fault in the algorithmic behavior; on the contrary, 
this is the desired convergence profile in canonical global optimization. 
At the same time, it precludes proper sampling about the optimum, and therefore no meaningful statistical learning may be carried out. 

\subsubsection{The Characteristic Update}
In order to generate ``sampling pressure'' that prevents stagnation and forces continuous landscape exploration about
the optimum, the CSA mechanism has to be replaced. 
Upon attempting to set a lower bound $\varepsilon$ on the global step-size $\sigma^{(g)}$, 
a rapid decrease of the covariance eigenvalues is observed, eliminating again proper sampling about the optimum (Eq.\ \ref{eq:pstepm}):
 \begin{equation}
  \displaystyle \sigma^{(g)}\xrightarrow[J\rightarrow 0]{} \varepsilon 
~~\Longrightarrow ~~\textrm{tr}\left(\mathbf{C}^{(g)}\right)  \xrightarrow[J\rightarrow 0]{} 0
~~\Longrightarrow ~~ \delta_p^{(g)}\xrightarrow[J\rightarrow 0]{} 0 .  
\end{equation}
In FOCAL, the CSA is replaced with a step-size update
that always forces a finite practical step-size $\delta_{p}^{(g)}$, regardless of the landscape location or mutation success.
FOCAL imposes the strongest spectrum-dependent pressure on the practical step-size, 
while counteracting the decrease of the covariance trace, $\textrm{tr}\left(\mathbf{C}^{(g)} \right)$, according to the following update:  
\begin{equation} \label{eq:focal} 
\displaystyle\boxed{
\sigma^{(g)} = \frac{\sigma_0} {\left( \lambda_{\min}^{(g)} \right)^{\alpha}}}
\end{equation}
Upon substituting Eq.\ \ref{eq:focal} into Eq.\ \ref{eq:pstepm} (with  
$\mu_{\textrm{eff}}=1$), the practical step now becomes
\begin{equation}
\delta_p^{(g)} = \sigma_0 \cdot \frac{\sqrt{{\textrm{tr}\left(\mathbf{C}^{(g)}\right)}}}
{\left( \lambda_{\min}^{(g)} \right)^{\alpha}}
\end{equation}
The scalar $\sigma_0$ is a constant user-specified forced step through search space, and the FOCAL learning power satisfies
$\alpha \leq 1/2$. 
According to this proposed rationale, new offspring are always required to make a step in all axes, and the collected
statistics allow refinement of the optimal sampling distribution required to maintain high fitness values. In order to stay at the top
of the landscape, FOCAL forces discovery of an optimal sampling distribution that generates mutations orthogonal to
yield-sensitive search directions.  

The operational mechanism of FOCAL and the role of the parameter $\alpha$ may be further understood 
upon consideration of the derived bounds for the practical step-size, $\delta_p^{(g)}$, when the global step-size is
updated subject to the FOCAL scheme (substituting Eq.\ \ref{eq:focal} into Eq.\ \ref{eq:bounds-pstepm}):
\begin{equation} \label{eq:practicalbound}
\displaystyle \boxed{\left( \lambda_{\min}^{(g)}\right)^{\frac{1}{2} -
\alpha} \lesssim \frac{\delta_p^{(g)}}{\sigma_0\sqrt{n}} \lesssim 
\sqrt{\textrm{cond}\left(\mathbf{C}^{(g)} \right)} \left(
\lambda_{\min}^{(g)}\right)^{\frac{1}{2} - \alpha}}
\end{equation}
The lower and upper bounds of the practical step-size $\delta_{p}^{(g)}$ are
dependent upon the covariance spectrum when $\alpha < 1/2$. The
lower bound corresponds to mutations taken entirely parallel to the most
yield-sensitive search direction (the smallest covariance eigenvalue, or alternatively
the largest Hessian eigenvalue), whereas the upper bound corresponds to mutations taken parallel 
to the least yield-sensitive search direction (largest covariance eigenvalue, smallest Hessian
eigenvalue).

In order to minimize the practical step in critical search
directions and therefore prevent significant yield deviations, the
proper covariance matrix must be learned.  As the
estimation of the covariance matrix improves (i.e.,
$\lambda_{\min} \rightarrow 0$), the occurrence of significant
yield deviations diminishes as mutations predominantly occur in
the search null-space directions.  
The spectral dependency, with $\alpha<1/2$, generates pressure
to learn the optimal covariance, since its removal will make the 
practical step-size identical in all directions. 
Furthermore, the condition number, 
\begin{equation}
\label{eq:cond}
 \displaystyle \textrm{cond}\left(\mathbf{C}^{(g)} \right) \equiv \frac{\lambda^{(g)}_{\max}}{\lambda^{(g)}_{\min}},
\end{equation} 
rapidly grows as the estimation of the covariance matrix improves, which may generate
significant steps outside of the quadratic
regime. Thus, the spectrum dependency also plays the role of
taming this rapid expansion of the practical step-size upper bound. The
optimal value of $\alpha$ is likely problem-dependent and
can affect the FOCAL learning rate.

\subsubsection{Inversion and Regularization}
Upon termination of the forced exploration about the global optimum,
the Hessian may then be recovered by inversion of the FOCAL-optimized covariance matrix, i.e., 
\begin{equation*}
\mathbf{H}^{(f)} = \left(\mathbf{C}^{(f)}\right)^{-1} = \mathbf{R}^{(f)} \left(\mathbf{\Lambda}^{(f)}\right)^{-1}\left(\mathbf{R}^{(f)}\right)^{T}, 
\end{equation*}
where the eigenvectors $\mathbf{R}^{(f)}$ are shared by both the covariance and Hessian, and as previously mentioned, the
Hessian spectrum is inversely proportional to the covariance spectrum.  However, because the optimal covariance matrix is rank
deficient, as evidenced by the ability to place mutations in the search null-space directions, this ill-conditioned matrix has 
first to be regularized prior to its inversion. It is worth noting that measurement uncertainty due to experimental noise may 
act as an implicit self-regularization procedure, and thus the covariance matrix only asymptotically approaches a singular form.  
In our proposed method, the Tikhonov filtering method is utilized for explicit regularization of the covariance spectrum 
\cite{regtutorial,NumRecipes} to yield the target Hessian spectrum,
\begin{equation}
\displaystyle h_i = \left(\lambda_i^{\textrm{reg}}\right)^{-1} = \frac{\lambda_i}{\lambda_i^2+\epsilon},~~~~~\epsilon\approx 10^{-7}.
\end{equation}

\subsubsection{Climbing the Landscape: Step-Size Alternatives}
As noted earlier, FOCAL critically relies on its driving search-engine to climb to the top of the landscape, 
where it can start launching its exploration -- i.e., it is essential for the underlying search-engine to succeed in order for FOCAL to function. 
As will become evident in Section \ref{sec:observation}, the proposed method is extremely efficient in QC experiments, which are the focus of this study.
It is possible that other search landscapes may not allow for an efficient climbing phase with the FOCAL 
routine, but would necessitate alternative step-size mechanisms, such as the CSA or others \cite{Akimoto}. 
In those cases, it is recommended to employ a DES search-engine for the climbing phase, 
and switch-on FOCAL once the search reaches the proximity of the global optimum, 
based upon a step-size criterion (e.g., once the global step-size reaches low values in the order  
of $\sigma^{(g)}\approx 10^{-4}$). 
The FOCAL method is summarized as Algorithm \ref{algo:FOCAL}.
\begin{algorithm}
\begin{algorithmic}[1]
\STATE \texttt{initDES()} 
\STATE \texttt{initParams}$\left(c_{cov},~\sigma_0,~\alpha\right)$
\STATE $g \leftarrow 0$
\REPEAT
\FOR {$k=1\ldots\lambda$}
     \STATE $\vec{x}^{(g+1)}_{k} \leftarrow \left<\vec{x}\right>_W^{(g)} + \sigma^{(g)} \cdot \vec{z}^{(g+1)}_k$,~~~~~$\vec{z}^{(g+1)}_k\sim \mathcal{N}\left(\vec{0},\mathbf{C}^{(g)} \right)$
     \STATE $f^{(g+1)}_{k}\leftarrow$ \texttt{evaluate} $\left(\vec{x}^{(g+1)}_{k}\right)$
\ENDFOR
\STATE $\left<\vec{x}\right>^{(g+1)}_W\leftarrow$ \texttt{select}$\left(\vec{f}^{(g+1)},~\vec{x}^{(g+1)}_{1\ldots\mu:\lambda}\right)$
\STATE $\vec{p}_c^{(g+1)}\leftarrow$ \texttt{updatePath} $\left(\vec{p}_c^{(g)},~\vec{z}^{(g+1)}_{1\ldots\mu:\lambda}\right)$ 
\STATE $\mathbf{C}^{(g+1)}\leftarrow$ \texttt{updateCovariance} $\left(c_{cov},~\mathbf{C}^{(g)},~\vec{p}_c^{(g+1)},~\vec{z}^{(g+1)}_{1\ldots\mu:\lambda}\right)$
\STATE $\mathbf{C}^{(g+1)} := \mathbf{R}^{(g+1)} \mathbf{\Lambda}^{(g+1)} \left(\mathbf{R}^{(g+1)}\right)^{T}$,
~~~ $\mathbf{\Lambda}^{(g+1)} = \textrm{diag} \left(\lambda_{\max}^{(g+1)},\ldots,\lambda_{\min}^{(g+1)}\right)$
\STATE $\boxed{\sigma^{(g+1)} \leftarrow \sigma_0 / \left( \lambda_{\min}^{(g+1)} \right)^{\alpha}}$ 
\STATE $g\leftarrow g+1$
\UNTIL stopping criterion is met
\STATE $\mathcal{C} \leftarrow$ \texttt{regularize}$\left(\mathbf{C}^{(g)}\right)$ \COMMENT{Tikhonov Filtering}
\STATE $\mathcal{H}\leftarrow \mathcal{C}^{-1}$
\RETURN $\mathcal{H}$
\end{algorithmic}
\caption{$(\mu_{W}, \lambda)$-FOCAL\label{algo:FOCAL}}
\end{algorithm}

\subsection{Preliminary Proof of Concept: A Noisy Model Landscape}
\begin{figure}
 \centering
\includegraphics[scale=0.75]{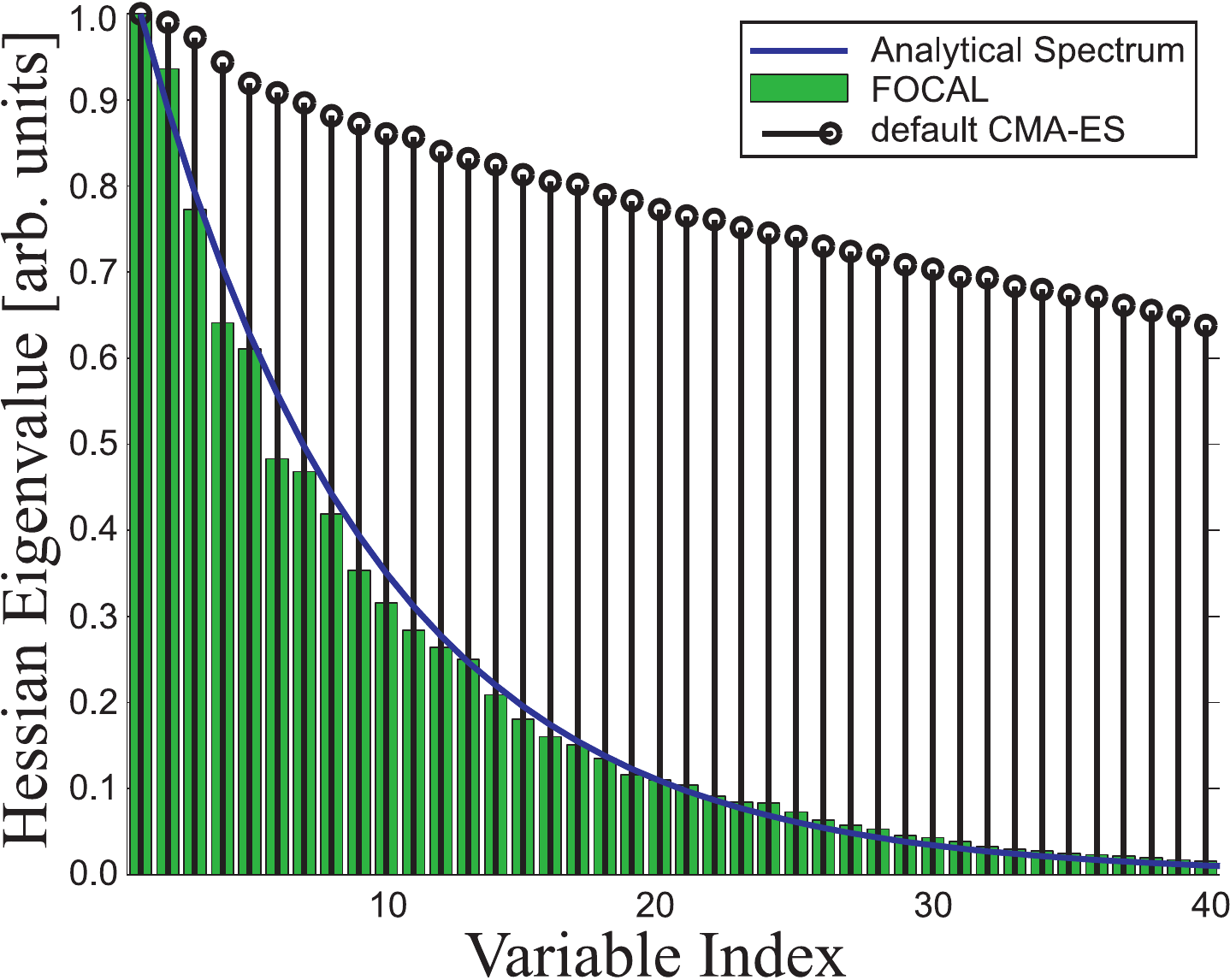}
\caption{Application of FOCAL to the separable ellipse (Eq.\ \ref{eq:ellipse}, $n=80$, $\xi=10^4$, $\epsilon_x=0.025$,
num/evals: $30,000$).
The recovered FOCAL spectrum (bars) accurately reproduces the analytical form (curve),
in contrast to the inaccurate recovery of the default CMA-ES (stems). 
The FOCAL parameters were set to $\sigma_0=0.075$, $c_{cov}=0.04$, $\alpha=0.10$.
\label{fig:ellipse}}
\end{figure}

As a preliminary demonstration of FOCAL operation, we consider a model landscape with noisy decision (input) variables,
known as the separable ellipse: 
\begin{equation}
\label{eq:ellipse}
\displaystyle f_{\textrm{e}}\left( \vec{x} \right) = \sum_{\imath=1}^n \xi^{\frac{\imath-1}{n-1}}\cdot \left(x_{\imath}+\mathcal{N}_{\imath}\left(0,\epsilon_x^2\right)\right)^2 \longrightarrow \min
\end{equation}
where $\xi$ is the condition number, set here to a high value of $10^4$.
The fact that this model landscape is a separable test-case does not constitute a limitation on our proof of concept,
as locating the global optimum is a necessary condition for FOCAL operation. By generating a challenging non-separable
instantiation of this problem, e.g., upon increasing the condition number and applying both translation and rotation (comparable
to \cite{Suganthan}), the success-rate of the employed DES kernel is simply reduced, and FOCAL's ability to recover landscape information 
is hampered respectively, as explained in the previous section. Since we would like to demonstrate the operation of FOCAL from
the bottom of the landscape, including the climbing-up phase, we will utilize the specified test-case.
\paragraph{Numerical Results} Fig.\ \ref{fig:ellipse} presents the outcome of FOCAL operating on the 80-dimensional
noisy separable ellipse with the def-CMA-ES search-engine. 
The numerical results clearly show that FOCAL obtains the correct Hessian information with satisfying accuracy. This outcome is in contrast to 
the default CMA-ES, which in this case fails to learn the correct eigenspectrum.

\section{Systems under Study: Quantum Control Experiments}
\label{sec:systems}
We demonstrate the operation of FOCAL on various experimental systems that require high-dimensional continuous optimization ($n=80$),
and at the same time, can benefit from Hessian determination about their global optima for sensitivity analysis, dimensionality reduction, mechanism
investigation, etc.
This section will present the systems under investigation in the field of experimental Quantum Control.
We will begin with a short introduction to this field, and especially with an overview on its practical optimization aspects. 
\subsection{Optimization in Quantum Control}
\begin{figure}
 \centering
\includegraphics[scale=0.5]{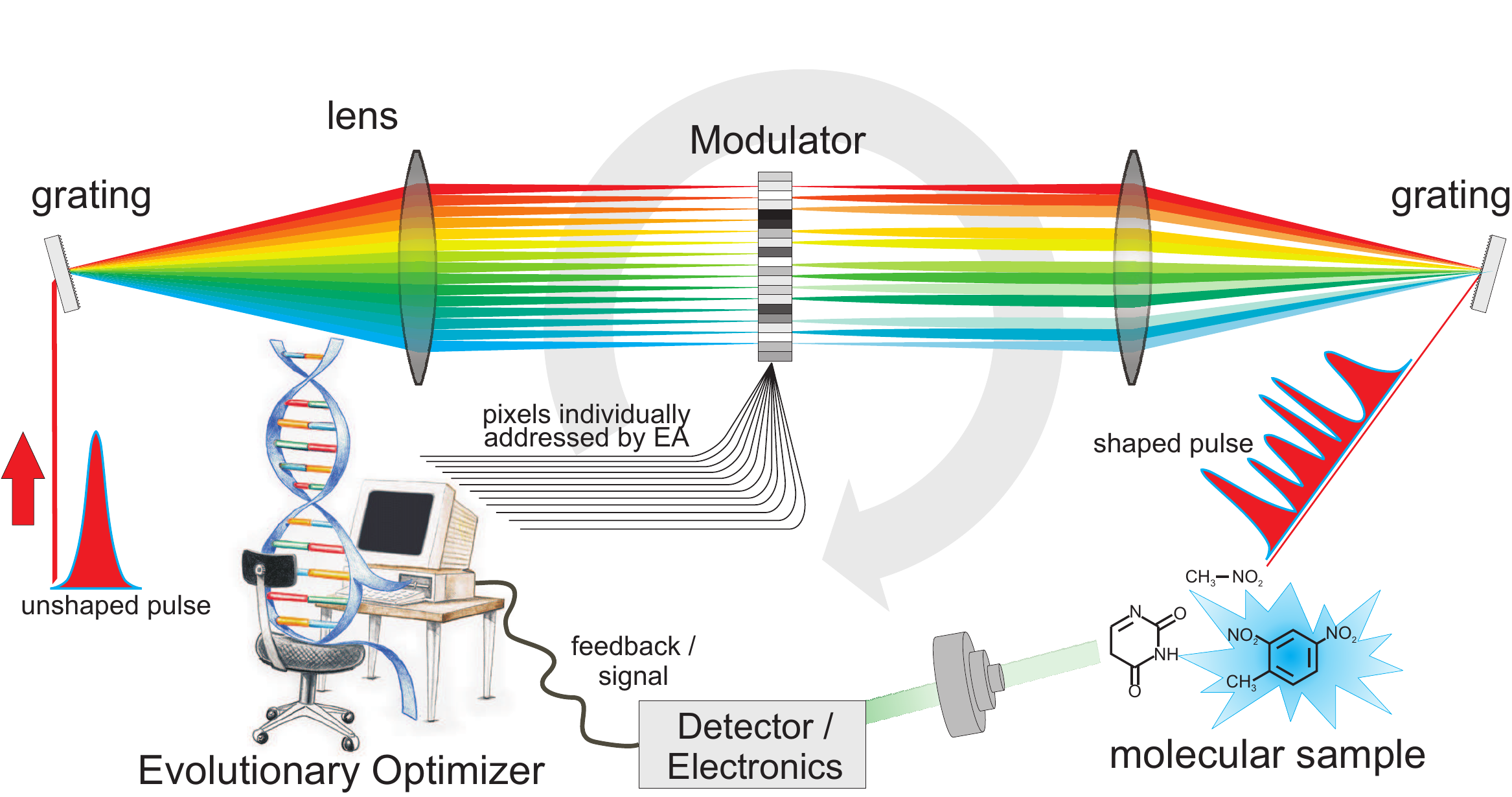}
\caption{The experimental Quantum Control learning loop. An unshaped pulse, approximated well by a Gaussian, is shaped by a pixel-based modulator and shined on 
a molecular sample. The decision variables addressed by the evolution strategy are the individual pixels that dictate the shaping. 
The measured signal reflecting the molecular response, which is provided by the detector, constitutes the feedback to the evolutionary optimizer. \label{fig:qce}}
\end{figure}

The primary laboratory operation in closed learning-loops within QC systems is the shaping of the temporal electric field, $E\left(t\right)$, 
which completely determines the dynamics of any controlled quantum process as dictated by the Schr\"odinger equation.
Determining $E\left( t \right)$ thus typically constitutes the optimization task. 
In practice, most QC experiments are carried out by means of spectral modulation, 
where the \emph{control function} consists of the spectral amplitude $A(\omega)$ and phase $\phi(\omega)$ functions,
which together construct the temporal electric field:
\begin{equation}
\label{eq:efield} 
E(t)=\mathbb{R} \left\{\int A(\omega)\exp(i\phi(\omega))
\exp(-i\omega t) \ d\omega \right\}.
\end{equation}
Most QC processes are very sensitive to the phase, and phase-only shaping is typically sufficient for attaining optimal
control. Our experiments include phase modulation only, where the spectral amplitude $A(\omega)$ is fixed. The latter function is
approximated well by a Gaussian and determines the bandwidth, or the minimal temporal pulse duration.
Shaping the field with phase-only modulation guarantees conservation of the pulse energy.
The spectral phase $\phi(\omega)$ is defined at $n$ frequencies
$\{\omega_j\}_{j=1}^n$ that are equally distributed across the
bandwidth of the spectrum. These $n$ values, $\{\phi(\omega_j)\}_{j=1}^n$, correspond to the $n$ pixels of the
pulse shaper and are the decision parameters to be optimized in the experimental learning loop:
\begin{equation} 
\label{eq:phase} 
\phi\left(\omega \right) = 
\left(\phi(\omega_1),\phi(\omega_2),...,\phi(\omega_n)\right). 
\end{equation}
The phase function is subject to periodic boundary conditions within $\left[0,2\pi\right]^n$. 
A practical note with regard to these boundary conditions will be given in the following section.
The pulse shaping process is implemented by a so-called Spatial Light Modulator (SLM), 
which is typically based on Liquid Crystal (LC) technology. This approach thus considers $n$ individual pixels, 
subject to rectangle-activation-functions, to construct the phase function $\phi(\omega)$.
Figure \ref{fig:qce} provides an illustration of an experimental QC learning loop.

\subsection{A Practical Note on Derandomization Upon Restricted Mutations}
In order to effectively address the periodic nature of the decision variables, a special treatment
enforcing the boundary conditions (see \cite{Hansen01completely} p.\ 179) can be incorporated into the optimization scheme. Overall,
there are three typical scenarios:
\begin{itemize}
\item Allowing unbounded free search (default optimization).
\item \emph{Rejecting} offspring which violate the boundary conditions $\left[ 0,2\pi\right]^n$.
\item Employing a \emph{wrapping operator}, i.e., $\phi(\omega)\leftarrow \phi(\omega) \textrm{mod}2\pi$, immediately after the mutation procedure.
\end{itemize}
The default unbounded search typically performs well in terms of attaining a good QC yield, but its learning rate is slow and overall performance inefficient.
At the same time, it obtains phase solutions that are hard to interpret, as they are stretched over large scales, far beyond the $2\pi$ domain.
Most importantly in our context, the learned correlations between the decision variables are not physical,
and it therefore cannot be employed in the FOCAL scheme.
As for the other two procedures, the \emph{rejection} concept clearly solves the boundary condition problem,
while potentially hampering the progress rate when discarding already evaluated candidate solutions; in contrast,
the \emph{wrapping} operation offers a simple solution which directly treats the physical issue, without losing evaluated offspring.
However, upon wrapping a newly generated offspring, it is critical within the derandomization framework to simultaneously update
the mutation vector which led to this candidate solution.
For instance, the \emph{a posteriori} mutation vector may be easily reconstructed within the CMA-ES in the following manner:
\begin{equation}
\label{eq:z_post}
\displaystyle \vec{z}_k^{post} = \left(\mathbf{R} \mathbf{\Lambda}^{1/2} \right)^{-1} \cdot \left( \frac{\vec{x}_k^{\textrm{mod}2\pi}-\left<\vec{x}\right>_W^{OLD}} {\sigma}\right).
\end{equation}
We thus adopt this proposed wrapping scheme.


\subsection{Implementation}
In the following sections we describe in detail our experimental results upon deploying FOCAL to the two QC experimental systems.
A commercial Ti:Sapphire femtosecond laser system generates amplified pulses centered at $780\textrm{nm}$ with a
bandwidth of $\Delta\lambda\sim36\textrm{nm}$, which gives pulses of duration $\Delta \tau \sim 30\textrm{fs}$ 
full-width-half-maximum (FWHM). These pulses are delivered to a pulse shaper with a
programmable 640 pixel SLM for phase-only modulation.
The search is conducted across an $n=80$ dimensional space of spectral phases (SLM pixels are tied together in groups of eight).

\subsection{Rank-Deficient Problem: Atomic Rubidium}
The first experimental system to which FOCAL is applied possesses a rank-deficient Hessian, 
and involves atomic Rubidium (Rb) \cite{dusi01}.
In this process, a shaped pulse induces atomic transitions in Rb,
while covering multiple resonant pathways, after which a radiative decay to the ground state 
produces visible fluorescence, which serves as a measure of the excited state population and forms the feedback signal.
This is a simple two-photon process, with a rank-deficient global maximum.
 
\paragraph{Laboratory Realization} Following phase-only shaping, the laser
pulses are focused into a $25\textrm{mm}$ vapor cell of atomic Rb maintained at $90^o$C.
The visible fluorescence is collected in a direction orthogonal to the incident beam. 

\subsection{Full-Rank Problem: Second Harmonic Generation (SHG)}
Second harmonic generation (SHG) or \emph{frequency doubling} is a
two-photon process in which an electric field interacts
nonlinearly with a material and generates an output photon with
the combined energy of two input photons.
The total energy of the output
light is proportional to the integrated squared intensity of the
primary pulse, as expected from a second-order nonlinear optical process.

The time-dependent profile of the laser field is given
in Eq.\ \ref{eq:efield}. The Total-SHG signal is then specified by:
\begin{equation}
\label{shg} SHG_t \equiv S_t = \int_{-\infty}^{+\infty}
I(t)^2~dt=\int_{-\infty}^{\infty}\left|E(t)\right|^4~dt,
\end{equation}
i.e., integration of the squared intensity. SHG is a common
test-case in the laboratory, as it constitutes a useful indication of the pulse intensity,
and its investigation contributes to the understanding of other processes. Moreover, its attractiveness also lies in its
complete mathematical formulation.
\paragraph{Optimization: Global Maximum}
In the context of global optimization, the maximization problem of the Total-SHG signal is often considered as a typical calibration task,
and has been previously reported \cite{QCE_GECCO08}. 
The Total-SHG signal $S_t$ is maximized by any linear phase function of frequency, and
in particular by a constant phase, i.e.,
\begin{equation}
\rm{argmax}_{\phi\left(\omega\right)}\left\{S_t\left(\phi\left(\omega\right)\right)\right\}\equiv
a\cdot \omega + b ~~~\forall a,b.
\end{equation}
Despite the simplicity of the optimal phase function, the fitness landscape has been shown to have a highly complex structure \cite{Roslund06}.

\paragraph{Laboratory Realization} The Total-SHG signal is observed by focusing amplified
pulses onto a 20 $\mu m$ type-I BBO crystal, and the actual signal is recorded with a photodiode and boxcar integrator.

%
%
%


\section{Practical Observation}
\label{sec:observation}

%

\subsection{Atomic Rubidium}
\begin{figure}
\centering
\includegraphics[scale=0.47]{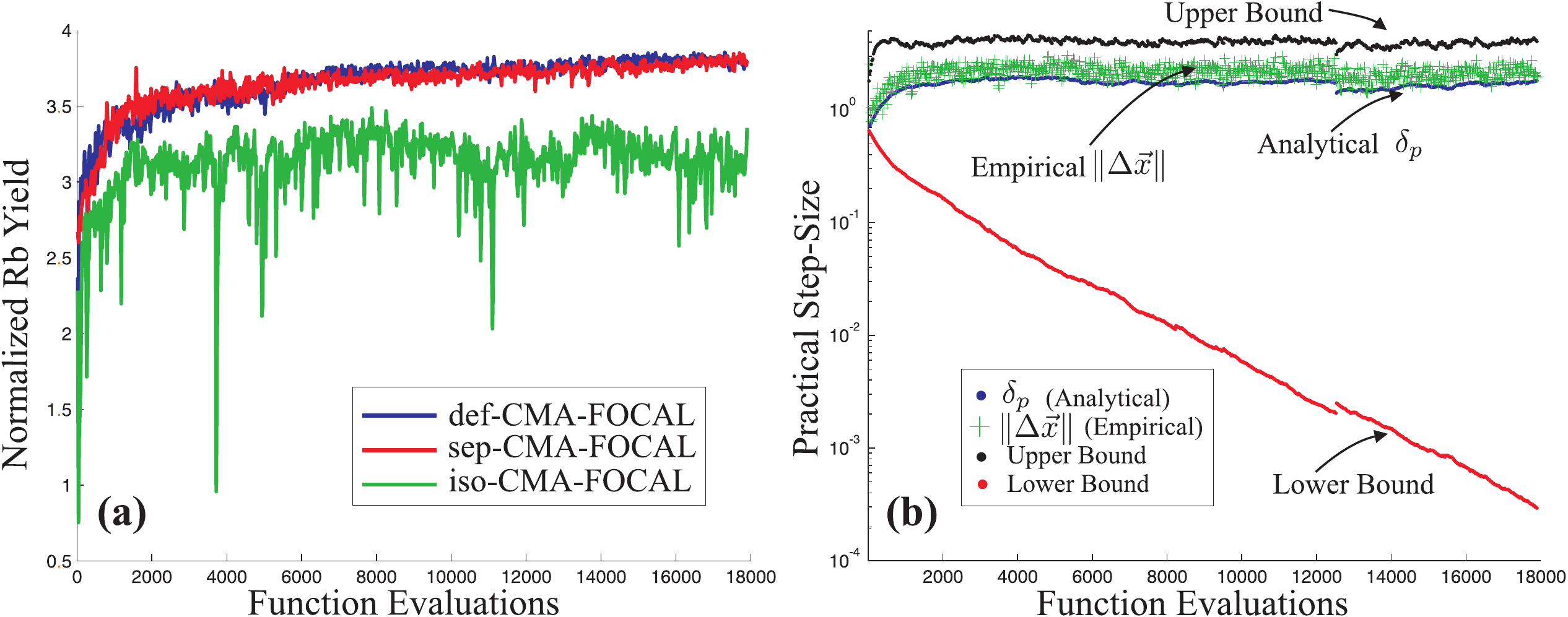}
\caption{Experimental FOCAL conducted on the atomic Rubidium system. 
Panel (a) displays the optimization yield curves (normalized to the transform-limited pulse) for the def-CMA-FOCAL, sep-CMA-FOCAL,
and the iso-CMA-FOCAL. While the curves of the def-CMA and sep-CMA practically merge, the iso-CMA routine fails to remain at the 
top of the landscape upon forcing a practical step within the FOCAL scheme.
Panel (b) displays post-facto calculations concerning the behavior of the practical step-size, namely, $\delta_p$ (Eq.\ \ref{eq:pstepm}), 
$\left\| \Delta \vec{x} \right\|$ (empirical, as the distance between consecutive search points), 
as well as the upper and lower bounds (Eq.\ \ref{eq:practicalbound}). The calculations correspond to the def-CMA-FOCAL run shown in panel (a). 
\label{fig:RbPS}}
\end{figure}

The forced step $\sigma_0$ is set to $7.5\%$ of $\left[0,2\pi\right]$, $\alpha=0.25$, and a $(10,20)$ strategy is employed.
Figure \ref{fig:RbPS}(a) presents the fitness curves of the three FOCAL routines
when applied to the optimization of the atomic Rb system.
The learning curves of the def-CMA and sep-CMA practically merge and  
rise above that of the iso-CMA following $\sim 1000$ experimental evaluations.  
Both search-engines continue to improve their fitness values and upon convergence
($\sim 7500$ experiments), fitness variations induced by either experimental noise or poor mutations diminish. 
Conversely, the iso-CMA optimization does not realize significant improvement after $\sim 1000$ experiments, 
and only achieves $\sim 85\%$ of the fitness obtained by either the def-CMA or the sep-CMA. 
Prominent signal fluctuations are visible and the iso-CMA also suffers from large fitness value drops. The
substandard and unstable performance of iso-CMA-driven FOCAL indicates
an inability to generate mutations solely in search directions
that are yield-insensitive (i.e., the search null-space) when subject to the current step-size mechanism. 

In order to supplement the evidence that FOCAL learns the optimal sampling distribution, it is insightful to additionally examine the post-facto practical step-sizes (shown in Figure \ref{fig:RbPS}(b)) taken across the fitness landscape for the def-CMA-driven FOCAL.
The calculations involve $\delta_p$ (Eq.\ \ref{eq:pstepm}), $\left\| \Delta \vec{x} \right\|$ 
(empirical, as the distance between consecutive search points), as well as the largest and smallest practical step-sizes 
(also referred to as the upper and lower bounds, respectively, corresponding to Eq.\ \ref{eq:practicalbound}).
After $\sim 1000$ experimental evaluations, the upper and lower bounds begin to meaningfully diverge. 
The practical step-size continues to lie near the upper bound for the remainder of the optimization, and the proximity 
of the experimental steps to this upper bound indicates that the generated mutations lie predominantly in a space spanned
by yield-insensitive search directions, which explains the minimal yield deviations seen in Figure \ref{fig:RbPS}(a).  
In essence, this observation constitutes the ultimate demonstration of the principle FOCAL rationale.

\begin{figure}
\centering
\includegraphics[scale=0.7]{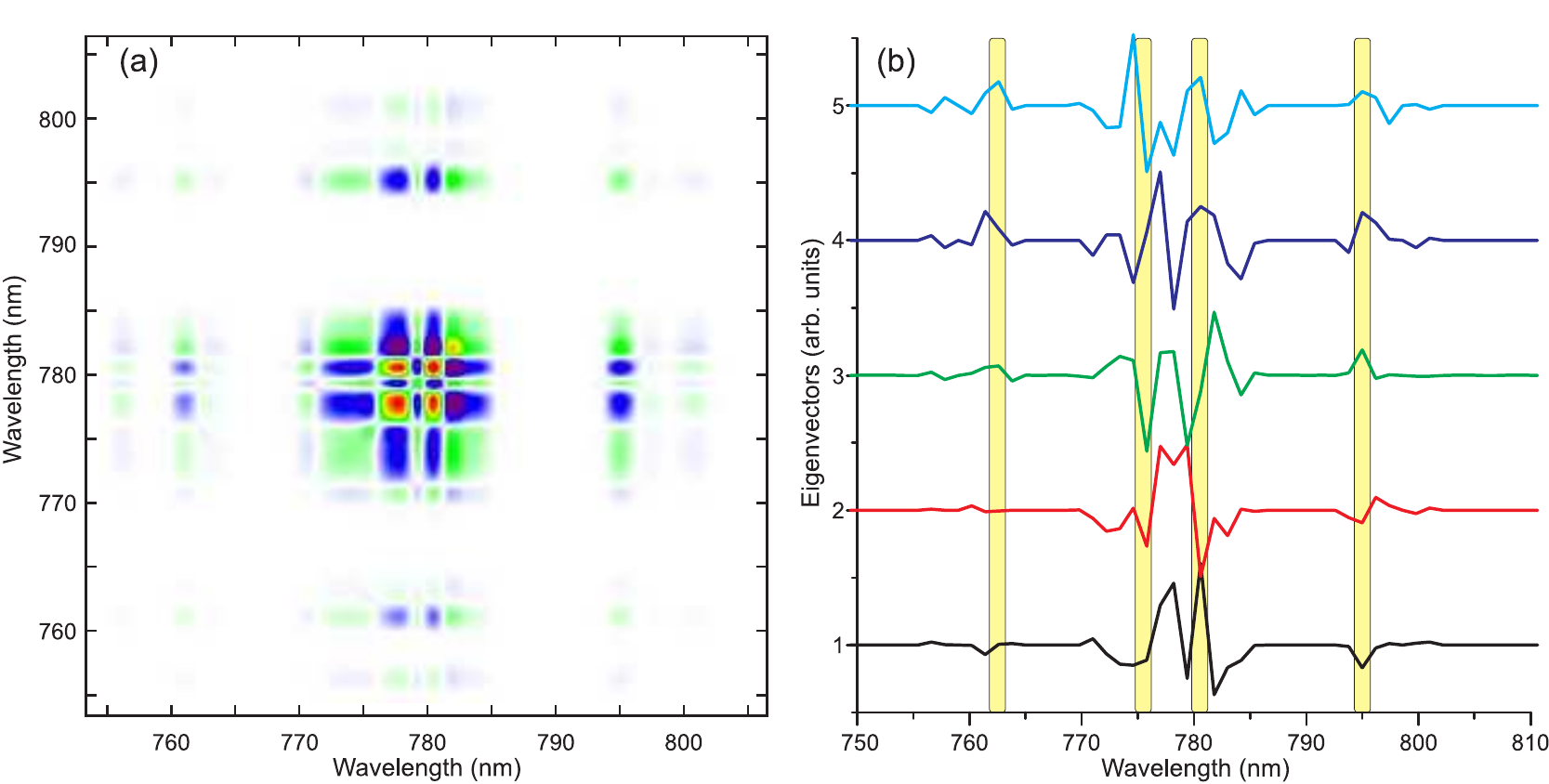}
\caption{Experimental Hessian for atomic Rubidium
obtained by FOCAL.  Panel (a) displays the Hessian matrix
retrieved from inverting the FOCAL-optimized Covariance matrix.
Panel (b) displays the eigenvectors corresponding to the five most
yield-sensitive search directions.  These eigenvectors exhibit
prominent structure surrounding the four resonant wavelengths of
rubidium, which are indicated with the colored bars.\label{fig:RbHessian}}
\end{figure}

The resultant Hessian recovered from the experiment is displayed in Figure \ref{fig:RbHessian}(a).
The Hessian illuminates individual couplings within Rb, 
and its eigendecomposition reveals the existence of only six important search directions
(matrix rank), which
explains the noise robustness observed in Figure \ref{fig:RbPS}(a)
as the majority of mutations lie within a null-space for the
quantum system.  In order to demonstrate the reliability of the
FOCAL-identified Hessian, the eigenvectors corresponding to the
five most yield-sensitive search directions (largest Hessian
eigenvalues) are shown in Figure \ref{fig:RbHessian}(b). As seen in the
figure, the recovered eigenvectors display prominent spectral
structure at the known four resonant wavelengths of Rb in addition
to the two-photon wavelength ($\sim$ 778nm). Yet, each eigenvector does not
simply correspond to a single resonant wavelength, but rather is a
complicated superposition of features across the entire spectral
region. These eigenvectors comprise a reduced-dimensional,
optimal control basis with which future optimizations or landscape
studies may be performed.
The eigenspectrum profile has also been successfully corroborated in two different manners:
Firstly, by conducting a na\"ive statistical Hessian measurement at the top of the landscape, involving 
$\approx$25,000 evaluations, and secondly, by applying PCA to the set of recorded search points at the 
final stage of the evolutionary search. 

\subsection{Total-SHG}
\begin{figure}
\centering
\includegraphics[scale=0.6]{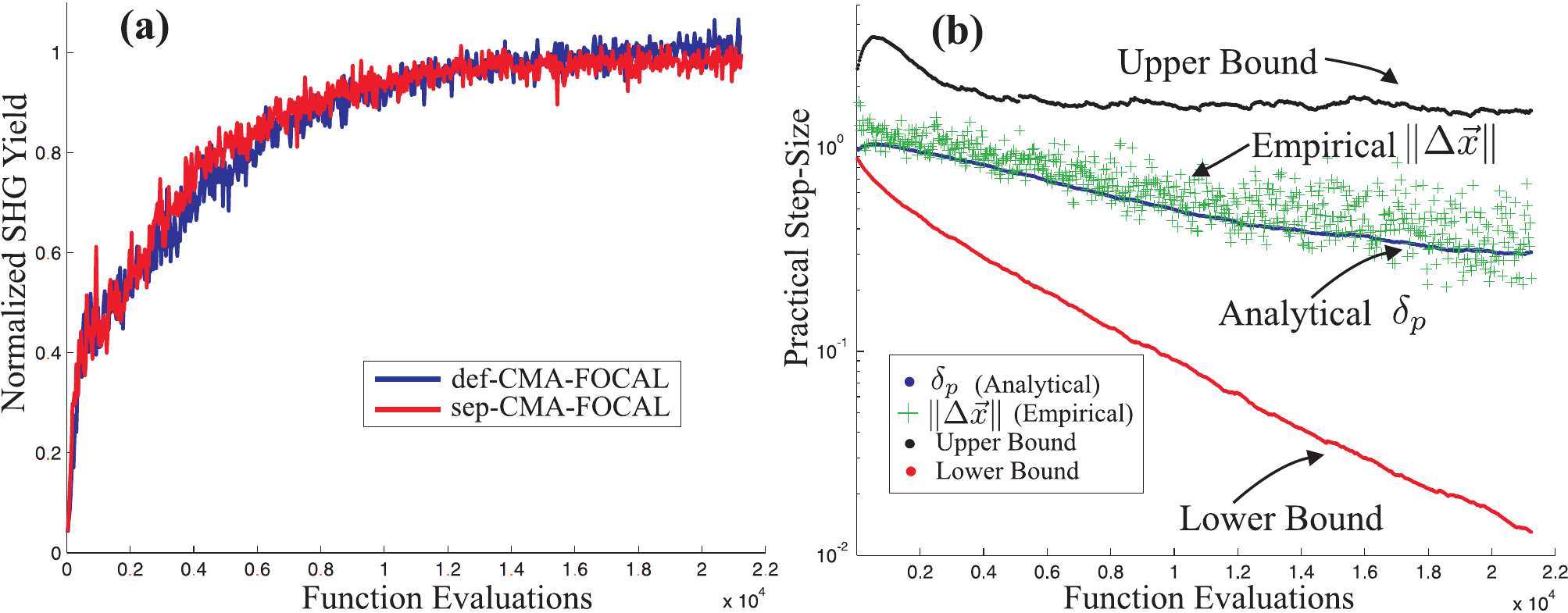}
\caption{Experimental FOCAL conducted on total-SHG. Panel (a) displays the optimization yield curves
(normalized to the transform-limited pulse) for the default CMA-ES (def-CMA) and the sep-CMA-ES routines,
when a forced step $\sigma_{0}$ is imposed. This plot confirms the successful convergence of both strategies.
Panel (b) displays \emph{post-facto} calculations concerning the behavior of the practical step-size, namely, $\delta_p$ (Eq.\ \ref{eq:pstepm}), 
$\left\| \Delta \vec{x} \right\|$ (empirical, as the distance between consecutive search points), 
as well as the upper and lower bounds (Eq.\ \ref{eq:practicalbound}). The calculations correspond to the def-CMA run shown in panel (a). 
Unlike the rank-deficient Rb problem, where the practical steps approach their upper bound (see Figure \ref{fig:RbPS}(b)),
the full-rank nature of the SHG problem dictates a proclivity toward the lower bound, as clearly shown here.
\label{fig:SHGPS}}
\end{figure}

The forced step $\sigma_0$ is set to $10\%$ of $\left[0,2\pi\right]$, $\alpha=0.1$, and a $(15,30)$ strategy is employed.
Figure \ref{fig:SHGPS}(a) presents the fitness curves of def-CMA and sep-CMA driven FOCAL routines when applied to the optimization
of the Total-SHG system. The two curves practically merge, as was also observed in the Rb optimization case. We omit details
regarding the iso-CMA-FOCAL, as its behavior was inferior, in an equivalent manner to the reported observation on the Rb system.
Furthermore, the calculated \emph{post-facto} practical steps are depicted in Figure \ref{fig:SHGPS}(b). 
Unlike the rank-deficient Rb problem, where the practical-steps approached their upper limit (see Figure \ref{fig:RbPS}(b)),
the full-rank nature of the SHG problem forces the practical-steps to approach their lower limit, as clearly shown.
This is an expected result, since the lack of a null space requires a more careful generation of mutations with a lower practical step.

The resultant Hessian recovered from this experiment yielded a spectrum which accurately matches
the analytical spectrum derived from theory (see Appendix), as shown in Figure \ref{fig:SHG_spectrum}.
This perfect match constitutes another corroboration of FOCAL's ability to successfully operate in this
high-dimensional regime and to extract complex information from experimental data.
In Figure \ref{fig:SHG_spectrum}, we again see the failure of the default CMA-ES to uncover 
the correct spectrum. 
\begin{figure}
\centering
\includegraphics[scale=0.75]{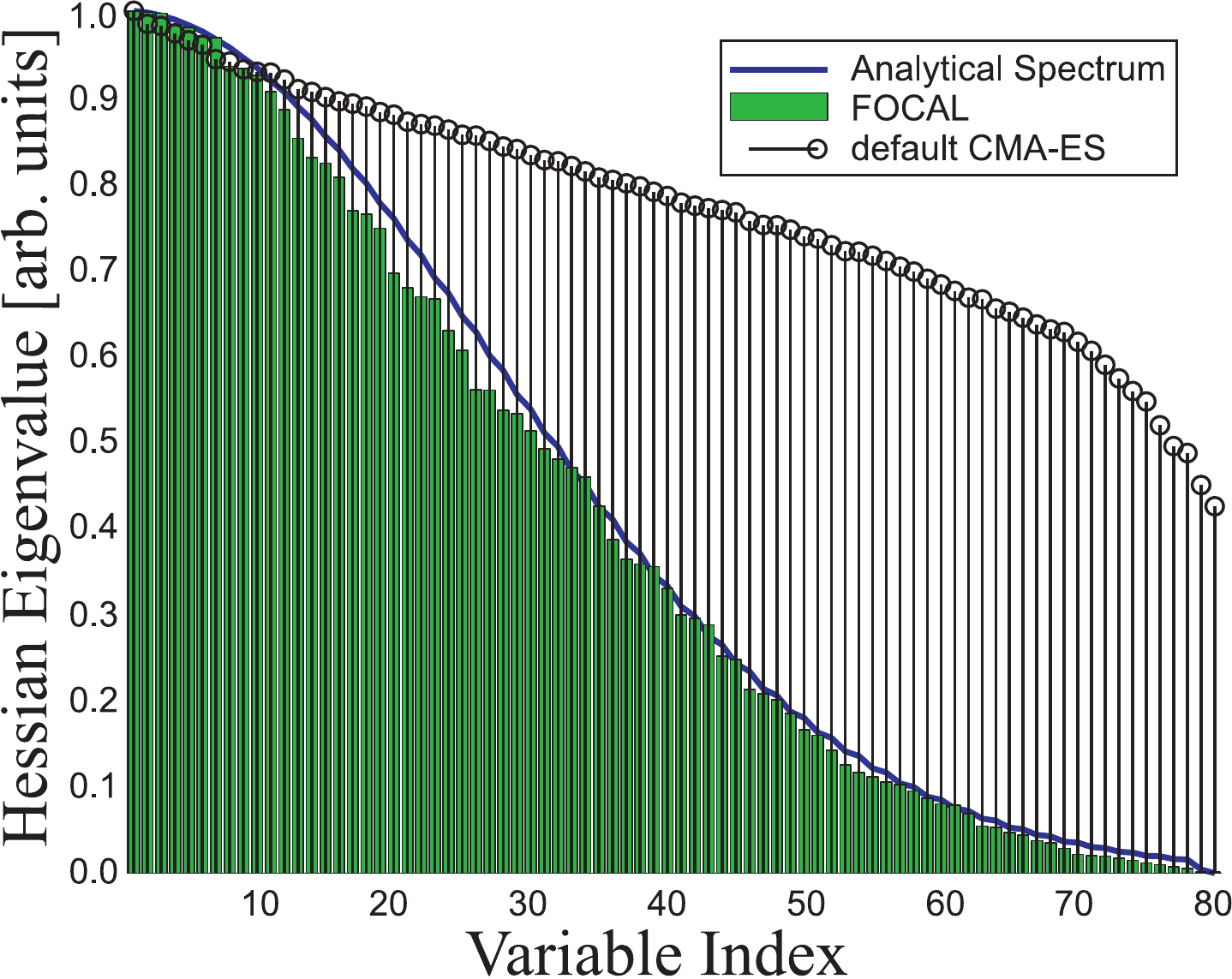}
\caption{Experimental SHG Hessian spectrum recovered by FOCAL (bars), which matches
the theoretical spectrum (curve) (for its derivation see the Appendix; note that the current scaling depicts the negative of the Hessian eigenvalues, as provided in Figure
\ref{fig-07}). The default CMA-ES (stems) fails to identify the correct eigenspectrum.
This problem is known to be non-separable and possesses a highly-complex, non-quadratic landscape \cite{Roslund06}.\label{fig:SHG_spectrum}}
\end{figure}



\subsection{Learning Rate: The Condition Number}
\begin{figure}
\centering
\includegraphics[scale=0.58]{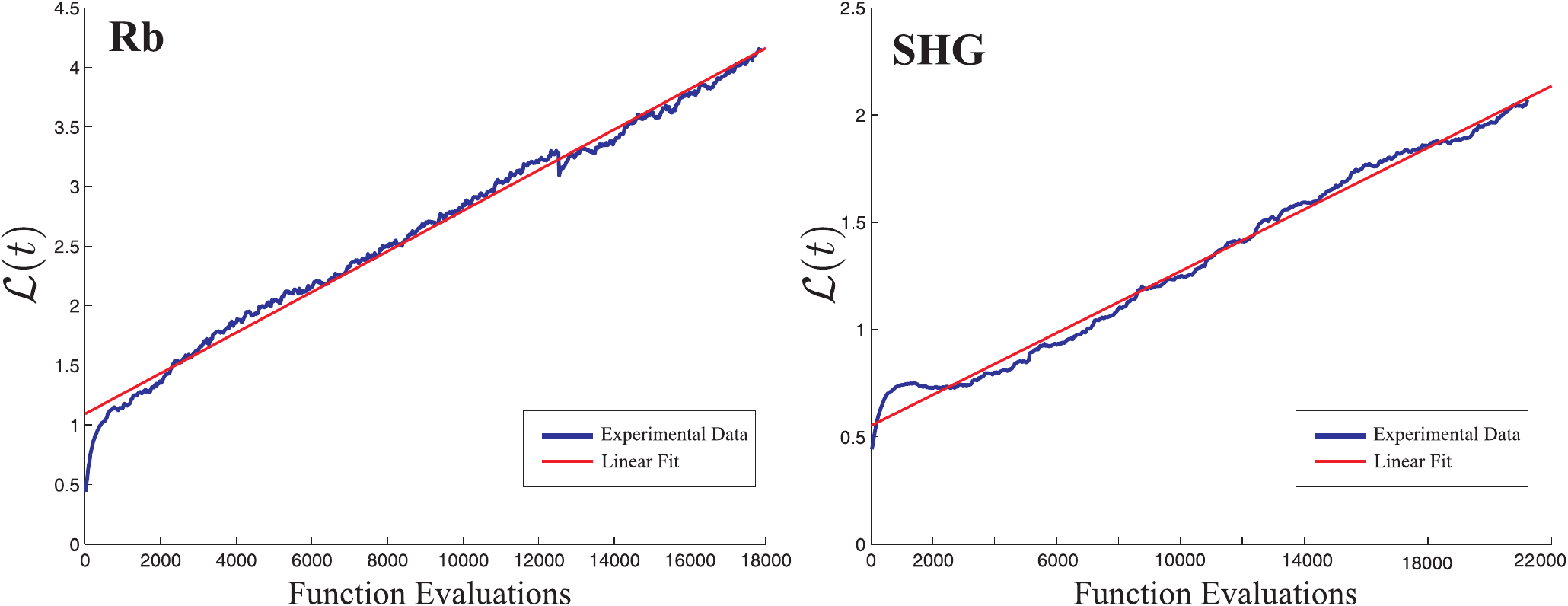}
\caption{The FOCAL learning rates, presented by the condition number as a function of the number of experimental trials. 
Linear fits are obtained for $\mathcal{L}(g)=\log_{10}\left[\sqrt{\textrm{cond}\left(\mathbf{C}^{(g)}\right)}\right]$, 
revealing \emph{exponential learning rates} for def-CMA-FOCAL operating on the two experimental QC systems:
[Left] Rb and [Right] Total-SHG.\label{fig:cond}}
\end{figure}

An important measure of algorithmic behavior is the learning rate of the target Hessian matrix.
This measure can be directly assessed by considering the condition number of the evolving covariance matrix (Eq.\ \ref{eq:cond}) as a function of
the number of experimental measurements (function evaluations). 
Figure \ref{fig:cond} presents the experimental data with regard to the learning rate. The results show
a linear increase of
\begin{equation*}
\mathcal{L}(g) = \log_{10}\left[\sqrt{\textrm{cond}\left(\mathbf{C}^{(g)}\right)}\right]
\end{equation*}
as a function of the number of experimental trials, for both the atomic Rb and Total-SHG systems.
Overall, this observation reveals a very attractive {\bf exponential learning rate} of the target matrix.
As a comparison, moment-based methods typically possess a rate proportional to the square-root of the number of experimental trials.
The complexity of the learning processes is reflected by the slopes of these curves: The rank-deficiency of the Rb system allows 
for a more rapid learning process in comparison to the Total-SHG problem with its full-rank Hessian. 

\subsection{Spectrum Evolution}
Investigation of the evolving spectra during FOCAL runs on the two systems reveals interesting information,
regarding both the search landscape as well as the FOCAL machinery. The evolving experimental spectra were captured in
movies, which are presented as online supplementary material \cite{Movie_focal-shg}. 
One observation is the rapid evolution of the two spectra as a function of the search stage, which confirms their locality.
In contrast to claims that the evolving covariance matrix within a self-adaptive ES is carrying
non-local information, we observe that the resultant spectrum within the FOCAL mechanism is locally formed only at the top of the landscape, i.e.,
it is not significantly contaminated by non-local information left over by the \emph{evolution path}, which constitutes a sliding average.
Another observation concerns the convergence of the spectra. In the atomic Rb system, the learning procedure exploits the rank-deficiency and first identifies the null space. 
Equivalently, in the Total-SHG case the procedure first identifies the directions with the smallest magnitude Hessian eigenvalues, i.e., it exploits a \emph{quasi-null} space for the construction of the spectrum. 
In the final convergence phase at the global maximum, it can be clearly seen for both systems that FOCAL is picking up the directions with the largest magnitude Hessian eigenvalues,
and gradually completes the spectra down to those with the lowest magnitude, in contrast to a possible scenario of simultaneous directional learning.
This is an expected result, which can be considered as a learning process driven by a greedy selection pressure striving for the least yield declines.

\subsection{Parameter Settings}
\label{sec:parameters}
In order to systematically characterize the role and the optimal values of the FOCAL parameters $c_{cov}$, $\alpha$, and $\sigma_0$,
we empirically studied FOCAL's behavior upon launching a series of simulations on model landscapes of varying ranks and dimensions.
Table \ref{tab:params} presents the recommended parameter values for different problem scenarios, and below are some
conclusions:
\begin{itemize}
\item FOCAL's behavior is not highly sensitive to the $\sigma_0$ value. We recommend setting it to $5\%$--$10\%$ of the coordinate interval.
\item $0 < \alpha < 1/2$ is mostly rank-dependent; the higher the rank, the lower its value should be set.
This is expected, as it controls the spectral pressure on FOCAL's machinery.
\item $1/n < c_{cov} < 1$ is both dimension- and rank-dependent. As the dimension and the rank increase, its value should be decreased.
\end{itemize}
\begin{table}
\begin{center}
\caption{Recommended FOCAL parameter settings, for $c_{cov}$ and $\alpha$, on a representative set of cases.\label{tab:params}}
\medskip
\begin{tabular}{|c||c|c||c|c|}
\hline
Dimension & \multicolumn{2}{c||}{Rank-Deficient} & \multicolumn{2}{c|}{Full-Rank} \\
\cline{2-5}
& $c_{cov}$ & $\alpha$ & $c_{cov}$ & $\alpha$\\
\hline
\hline
$n=30$ & 0.10 & 0.25 & 0.08 & 0.19 \\
\hline
$n=50$ & 0.08 & 0.22 & 0.06 & 0.15 \\
\hline
$n=80$ & 0.07 & 0.20 & 0.04 & 0.10 \\
\hline
\end{tabular}
\end{center}
\end{table}
Note that the table is practically symmetric with respect to the role played by either the rank or the dimension:
high-dimensional problems with rank-deficiency are empirically equivalent to low-dimensional full-rank problems.

\section{Discussion}
\label{sec:discussion}
We have introduced a new technique, entitled FOCAL, for extraction of second-order landscape information at the global basin of attraction,
which is based upon the CMA-ES search-engine. We derived fundamental principles for the inverse relation between the statistically learned ES
covariance matrix and the Hessian of the landscape. An algorithm was formulated, which \emph{forces} the ES-kernel to explore the top of the landscape
toward the end of an optimization routine efficiently determining the Hessian.
We described the method in detail, while deriving bounds for the characteristic adaptive strategy parameters, which allow for better
mechanism understanding. In summary, we claim that the inability of conventional ES to learn the correct covariance matrix under certain conditions most likely
lies in the global step-size behavior. Consequently, the primary contribution of FOCAL is the introduction of an alternative step-size mechanism 
specifically targeting Hessian determination. 
Within this scheme, the spectrum-dependency of FOCAL's step-size update seems to play a crucial role in the method's 
capability to learn the nature of the top of the landscape, 
especially when operating under conditions of large condition numbers ($\xi\gtrsim 10^4$).
This method also introduced several user-defined parameters along with  
proposed rules of thumb for setting those parameters.
Upon utilizing the proposed values, FOCAL was demonstrated to perform extremely well over a broad experimental platform.
Further study of the optimal FOCAL parameter settings may constitute an important direction for future research.

Following a demonstration of a practical scenario in which the default CMA-ES does not learn the correct Hessian spectrum of a basic model landscape, 
the newly proposed FOCAL method was shown to easily obtain the known Hessian.
It was then applied to the experimental platform of QC systems. We reported on the laboratory success of FOCAL to 
extract Hessian forms of several systems. Those experimental results reliably matched the Hessian forms expected from quantum mechanical theory,
and therefore constituted an experimental corroboration for the high fidelity of the FOCAL procedure.
The learning rate of FOCAL was observed to be exponential in terms of algorithmic iterations (or function evaluations), 
introducing an additional attractive feature of this novel technique.

This study offers an additional important observation with regard to the nature of the landscapes of the investigated QC systems.  
Simple intuition might suggest that the ease of optimizing QC systems is due to their possessing spherical landscapes. 
This study clearly refutes this behavior and explicitly demonstrates the non-spherical nature of the investigated QC landscapes.

There are two primary directions for future research. One is the investigation of possible spectrum compression by means of a first-order DES 
(a strategy employing a diagonalized covariance matrix, e.g., sep-CMA-ES).
Since the deployment of such a strategy was demonstrated to be successful within the FOCAL framework 
in terms of convergence, it would be interesting to rigorously study the nature of the attained compressed spectra, and ideally
draw an analytical link to the actual landscape information.
Another direction would be the characterization of the problem conditions under which conventional ES succeed or fail to learn the correct covariance matrix reflective of the search landscape. 
\subsection*{Acknowledgments}
The authors would like to thank Monte Lunacek and Tak-San Ho for the valuable discussions.
The authors acknowledge support from NSF and ARO.

\section*{Appendix: Derivation of the SHG Hessian}
\input{shgHessian}

\input{oshir_focal.bbl}

\end{document}

%% file: shgHessian.tex
\begin{figure}
 \centering
\includegraphics[scale=0.8]{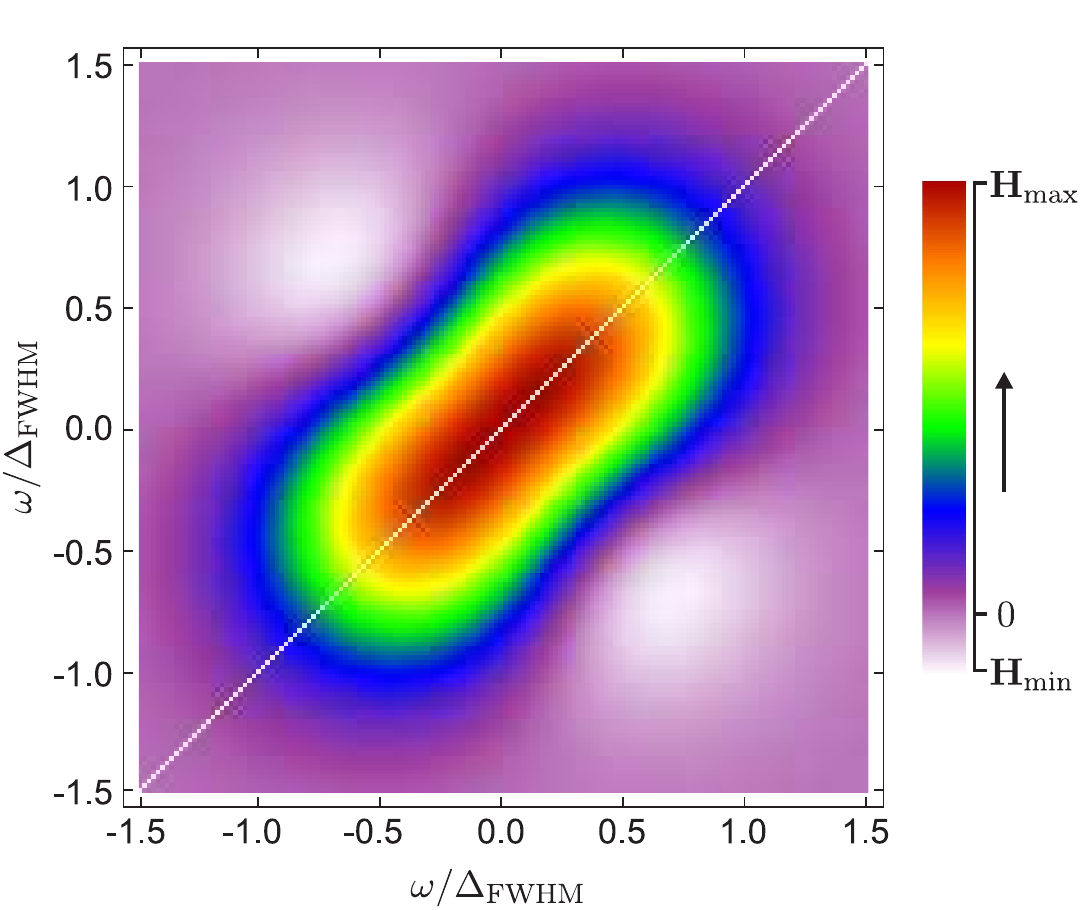}
\caption{Theoretical SHG Hessian matrix for a Gaussian spectral 
intensity with a bandwidth of $\Delta_{\textrm{FWHM}}$. \label{fig-07}}
\end{figure}

The Hessian for Second Harmonic Generation (SHG) is given by

\begin{equation}
\mathbf{H}(\omega',\omega'') = \left. \frac{\delta^2
J_{\textrm{SHG}}}{\delta \phi(\omega')~\delta \phi(\omega'')}
\right|_{\phi^{*}(\omega) = 0},
\end{equation}
where $\phi(\omega)$ is the spectral phase applied to the pulse
shaper.  The SHG signal $J_{\textrm{SHG}}$ is provided by

\begin{equation}
J_{\textrm{SHG}} = \int_{-\infty}^{\infty} \left| E_{2}(\omega)
\right|^2 d \omega,
\end{equation}
where $E_{2}(\omega) = \int_{-\infty}^{\infty} E(\omega') E(\omega
- \omega') d \omega'$ and the complex spectral field $E(\omega) =
A(\omega) \exp[i \phi(\omega)]$ is given in terms of the laser
spectral amplitude $A(\omega)$ and applied phase $\phi(\omega)$.
The spectral amplitude $A(\omega)$ is taken as real, positive, and
related to the experimentally measured spectral intensity
$|E(\omega)|^2$ through $|E(\omega)|^2 = A(\omega)^2$. The
second-order functional derivative of $J_{\textrm{SHG}}$ may then
be written as

\begin{equation} \label{eq:hessian1}
\frac{\delta^2 J_{\textrm{SHG}}}{\delta \phi(\omega')~\delta
\phi(\omega'')} = \int_{-\infty}^{\infty} d \omega \left(
\frac{\delta^2 E_{2}(\omega)}{\delta \phi(\omega')~\delta
\phi(\omega'')} E_2^{*}(\omega) + \frac{\delta
E_{2}(\omega)}{\delta \phi(\omega')}~\frac{\delta
E_{2}^{*}(\omega)}{\delta \phi(\omega'')} + c.c. \right).
\end{equation}
Variational analysis is utilized to derive the relevant functional
derivatives of the complex spectral field:

\begin{align*}
\frac{\delta E_{2}(\omega)}{\delta \phi(\omega')} &= 2 i
E(\omega')
E(\omega - \omega') \\
\frac{\delta^2 E_{2}(\omega)}{\delta \phi(\omega')~\delta
\phi(\omega'')} &= -2 E(\omega')E(\omega-\omega') \left[
\delta(\omega' - \omega'') + \delta(\omega - \omega' - \omega'')
\right].
\end{align*}
Substitution of these two relations into Eq.~\ref{eq:hessian1} and
evaluation with the optimal phase ($\phi^{*}(\omega) = 0$) finally
leads to a SHG Hessian of the form

\begin{align} \label{eq:hessian2}
\mathbf{H}(\omega',\omega'') = &-4~A(\omega') \delta (\omega' -
\omega'') \iint\limits_{-\infty}^{\hspace{10pt} \infty} d \omega d
z A(z) A(\omega -
\omega') A(\omega - z) \nonumber \\
&-4~A(\omega') A(\omega'') \int\limits_{-\infty}^{\hspace{10pt}
\infty} d\omega
A(\omega)A(\omega'+\omega''-\omega) \nonumber \\
&+8~A(\omega') A(\omega'') \int\limits_{-\infty}^{\hspace{10pt}
\infty} d\omega A(\omega-\omega') A(\omega-\omega'').
\end{align}
The Hessian may be cast into a more tractable expression by considering an
experimentally relevant spectral amplitude of the form $A(\omega)
= \exp[-2 \ln 2 (\omega^2 / \Delta_{\textrm{FWHM}}^2)]$, where
$\Delta_{\textrm{FWHM}}$ is the spectral intensity FWHM. Following
this assumption, the integrals of Eq.~\ref{eq:hessian2} may be
analytically evaluated, and the Hessian becomes

\begin{align} \label{eq:hessian3}
\mathbf{H}(\omega',\omega'') =
&-4~\frac{\Delta_{\textrm{FWHM}}^2}{\sqrt{3}}
\frac{\pi}{\ln 2}~A(\omega') \exp[-\frac{2}{3} (\omega'^2 / \Delta_{\textrm{FWHM}}^2)]~\delta(\omega' - \omega'') \nonumber \\
&-4~\Delta_{\textrm{FWHM}} \sqrt{\frac{\pi}{\ln
2}}~A(\omega')A(\omega'')
\exp[-(\omega'-\omega'')^2 / \Delta_{\textrm{FWHM}}^2] \nonumber \\
&+8~\Delta_{\textrm{FWHM}} \sqrt{\frac{\pi}{\ln
2}}~A(\omega')A(\omega'') \exp[-(\omega'+\omega'')^2 /
\Delta_{\textrm{FWHM}}^2].
\end{align}
This Hessian is evaluated over the frequency domain, 
\begin{equation*}
\omega \in \left[ -1.5 \cdot \Delta_{\textrm{FWHM}}, 1.5 \cdot
\Delta_{\textrm{FWHM}} \right], 
\end{equation*}
which corresponds to the
experimental width of the pulse shaper pixel array (the optics of
the pulse shaper are designed so that the pixel array width is $3
\cdot \Delta_{\textrm{FWHM}}$). The resulting Hessian is depicted
in Figure~\ref{fig-07}. It displays both positive (diagonal) and negative
(off-diagonal) correlations amongst frequency components.

%% file: oshir_focal.bbl
\begin{thebibliography}{10}

\bibitem{Beyer-Schwefel}
Beyer, H.G., Schwefel, H.P.:
\newblock {E}volution {S}trategies \-a {C}omprehensive {I}ntroduction.
\newblock {Natural Computing: An International Journal} \textbf{1}(1) (2002)
  3--52

\bibitem{Hansen01completely}
Hansen, N., Ostermeier, A.:
\newblock {C}ompletely {D}erandomized {S}elf-{A}daptation in {E}volution
  {S}trategies.
\newblock Evolutionary Computation \textbf{9}(2) (2001)  159--195

\bibitem{HansenDR1}
Ostermeier, A., Gawelczyk, A., Hansen, N.:
\newblock {A Derandomized Approach to Self Adaptation of Evolution Strategies}.
\newblock Technical Report TR-93-003, TU Berlin (1993)

\bibitem{HansenDR2}
Ostermeier, A., Gawelczyk, A., Hansen, N.:
\newblock {Step-Size Adaptation Based on Non-Local Use of Selection
  Information}.
\newblock In: Parallel Problem Solving from Nature - PPSN III. Volume 866 of
  Lecture Notes in Computer Science., Springer (1994)  189--198

\bibitem{HansenDR3}
Hansen, N., Ostermeier, A., Gawelczyk, A.:
\newblock {On the Adaptation of Arbitrary Normal Mutation Distributions in
  Evolution Strategies: The Generating Set Adaptation}.
\newblock In: Proceedings of the Sixth International Conference on Genetic
  Algorithms (ICGA6), San Francisco, CA, Morgan Kaufmann (1995)  57--64

\bibitem{HansenDR4}
Hansen, N., Ostermeier, A.:
\newblock {Adapting Arbitrary Normal Mutation Distributions in Evolution
  Strategies: the Covariance Matrix Adaptation}.
\newblock In: Proceedings of the 1996 {IEEE} International Conference on
  Evolutionary Computation, {P}iscataway, {NJ}, {IEEE} (1996)  312--317

\bibitem{hansencmamultimodal}
Hansen, N., Kern, S.:
\newblock Evaluating the {CMA} {E}volution {S}trategy on {M}ultimodal {T}est
  {F}unctions.
\newblock In: Parallel Problem Solving from Nature - PPSN V. Volume 1498 of
  Lecture Notes in Computer Science., Amsterdam, Springer (1998)  282--291

\bibitem{HansenCEC2005b}
Auger, A., Hansen, N.:
\newblock {Performance Evaluation of an Advanced Local Search Evolutionary
  Algorithm}.
\newblock In: Proceedings of the 2005 Congress on Evolutionary Computation
  CEC-2005, Piscataway, NJ, USA, IEEE Press (2005)  1777--1784

\bibitem{HansenGECCO09}
Hansen, N.:
\newblock Benchmarking a bi-population cma-es on the bbob-2009 noisy testbed.
\newblock In: GECCO '09: Proceedings of the 11th annual conference companion on
  Genetic and evolutionary computation conference, New York, NY, USA, ACM
  (2009)  2397--2402

\bibitem{CMAapplications}
Hansen, N.:
\newblock {References to CMA-ES Applications}.
\newblock \url{http://www.lri.fr/~hansen/cmaapplications.pdf} (2009)

\bibitem{CMA-MO}
Igel, C., Hansen, N., Roth, S.:
\newblock {Covariance Matrix Adaptation for Multi-objective Optimization}.
\newblock Evolutionary Computation \textbf{15}(1) (2007)  1--28

\bibitem{hansen2009tec}
Hansen, N., Niederberger, S., Guzzella, L., Koumoutsakos, P.:
\newblock {A Method for Handling Uncertainty in Evolutionary Optimization with
  an Application to Feedback Control of Combustion}.
\newblock IEEE Transactions on Evolutionary Computation \textbf{13}(1) (2009)
  180--197

\bibitem{Shir-NACO08}
Shir, O.M., B{\"a}ck, T.:
\newblock {N}iching with {D}erandomized {E}volution {S}trategies in
  {A}rtificial and {R}eal-{W}orld {L}andscapes.
\newblock {N}atural {C}omputing: {A}n {I}nternational {J}ournal \textbf{8}(1)
  (2009)  171--196

\bibitem{Shir-SA_ECJ}
Shir, O.M., Emmerich, M., B{\"a}ck, T.:
\newblock {A}daptive {N}iche-{R}adii and {N}iche-{S}hapes {A}pproaches for
  {N}iching with the {CMA-ES}.
\newblock Evolutionary Computation \textbf{18}(1) (2010)  97--126

\bibitem{BeyerCMA}
Beyer, H.G., Sendhoff, B.:
\newblock {Covariance Matrix Adaptation Revisited -- The CMSA Evolution
  Strategy}.
\newblock In: Parallel Problem Solving from Nature - PPSN X. Volume 5199 of
  Lecture Notes in Computer Science., Springer (2008)  123--132

\bibitem{Akimoto}
Akimoto, Y., Sakuma, J., Ono, I., Kobayashi, S.:
\newblock {Functionally Specialized CMA-ES: A Modification of CMA-ES based on
  the Specialization of the Functions of Covariance Matrix Adaptation and Step
  Size Adaptation}.
\newblock In: GECCO '08: Proceedings of the 10th annual conference on Genetic
  and evolutionary computation, New York, NY, USA, ACM (2008)  479--486

\bibitem{CMAES-Tutorial}
Hansen, N.:
\newblock {The CMA Evolution Strategy: A Tutorial}.
\newblock \url{http://www.lri.fr/~hansen/cmatutorial.pdf} (2010)

\bibitem{Hersch93}
{Warren}, W.S., {Rabitz}, H., {Dahleh}, M.:
\newblock {Coherent Control of Quantum Dynamics: The Dream Is Alive}.
\newblock Science \textbf{259} (1993)  1581--1589

\bibitem{Hersch00}
Rabitz, H., de~Vivie-Riedle, R., Motzkus, M., Kompa, K.:
\newblock {Whither the Future of Controlling Quantum Phenomena?}
\newblock Science \textbf{288} (2000)  824--828

\bibitem{Gerber07}
Nuernberger, P., Vogt, G., Brixner, T., Gerber, G.:
\newblock {Femtosecond Quantum Control of Molecular Dynamics in the Condensed
  Phase}.
\newblock Phys Chem Chem Phys. \textbf{9}(20) (2007)  2470--2497

\bibitem{Hersch88}
Peirce, A.P., Dahleh, M.A., Rabitz, H.:
\newblock {Optimal Control of Quantum-Mechanical Systems: Existence, Numerical
  Approximation, and Applications}.
\newblock Phys. Rev. A \textbf{37}(12) (1988)

\bibitem{Hersch92}
Judson, R.S., Rabitz, H.:
\newblock {Teaching Lasers to Control Molecules}.
\newblock Phys. Rev. Lett. \textbf{68}(10) (1992)  1500--1503

\bibitem{baumert97}
Baumert, T., Brixner, T., Seyfried, V., Strehle, M., Gerber, G.:
\newblock Femtosecond pulse shaping by an evolutionary algorithm with feedback.
\newblock Applied Physics B \textbf{65} (1997)  779--782

\bibitem{zemo01}
Zeidler, D., Frey, S., Kompa, K.L., Motzkus, M.:
\newblock {Evolutionary Algorithms and their Application to Optimal Control
  Studies}.
\newblock Phys. Rev. A \textbf{64}(2) (2001)  023420

\bibitem{pearson01}
Pearson, B., White, J., Weinacht, T., Bucksbaum, P.:
\newblock Coherent control using adaptive learning algorithms.
\newblock Physical Review A \textbf{63}(6) (2001)  063412

\bibitem{LabES}
Roslund, J., Shir, O.M., B\"ack, T., Rabitz, H.:
\newblock {Accelerated Optimization and Automated Discovery with Covariance
  Matrix Adaptation for Experimental Quantum Control}.
\newblock Physical Review A (Atomic, Molecular, and Optical Physics)
  \textbf{80}(4) (2009)  043415

\bibitem{Hersch06}
Ho, T.S., Rabitz, H.:
\newblock {Why do Effective Quantum Controls Appear Easy to Find?}
\newblock Journal of Photochemistry and Photobiology A: Chemistry
  \textbf{180}(3) (2006)

\bibitem{Raj07}
Chakrabarti, R., Rabitz, H.:
\newblock {Quantum Control Landscapes}.
\newblock International Reviews in Physical Chemistry \textbf{26}(4) (2007)
  671--735

\bibitem{Christian-Elsevier}
{Siedschlag}, C., {Shir}, O.M., {B{\"a}ck}, T., {Vrakking}, M.J.J.:
\newblock {Evolutionary Algorithms in the Optimization of Dynamic Molecular
  Alignment}.
\newblock Optics Communications \textbf{264} (2006)  511--518

\bibitem{SHIR_CEC06}
Shir, O.M., Siedschlag, C., B{\"a}ck, T., Vrakking, M.J.:
\newblock {Evolutionary Algorithms in the Optimization of Dynamic Molecular
  Alignment}.
\newblock In: Proceedings of the 2006 IEEE World Congress on Computational
  Intelligence, IEEE Computational Intelligence Society (2006)  9817--9824

\bibitem{Shir-JPhysB}
Shir, O.M., Beltrani, V., B{\"a}ck, T., Rabitz, H., Vrakking, M.J.:
\newblock {On the Diversity of Multiple Optimal Controls for Quantum Systems}.
\newblock Journal of Physics B: Atomic, Molecular and Optical Physics
  \textbf{41}(7) (2008)  074021

\bibitem{QCE_GECCO08}
Shir, O.M., Roslund, J., B{\"a}ck, T., Rabitz, H.:
\newblock {Performance Analysis of Derandomized Evolution Strategies in Quantum
  Control Experiments}.
\newblock In: Proceedings of the Genetic and Evolutionary Computation
  Conference, GECCO-2008, New York, NY, USA, ACM Press (2008)  519--526

\bibitem{LarsJennifer}
Fanciulli, R., Willmes, L., Savolainen, J., van~der Walle, P., B\"ack, T.,
  Herek, J.L.:
\newblock {Evolution Strategies for Laser Pulse Compression}.
\newblock In: Proceedings of the International Conference Evolution
  Artificielle. Volume 4926 of Lecture Notes in Computer Science., Springer
  (2008)  219--230

\bibitem{BartelsCMA}
Wilson, J.W., Schlup, P., Lunacek, M., Whitley, D., Bartels, R.A.:
\newblock Calibration of liquid crystal ultrafast pulse shaper with common-path
  spectral interferometry and application to coherent control with a covariance
  matrix adaptation evolutionary strategy.
\newblock Review of Scientific Instruments \textbf{79}(3) (2008)  033103+

\bibitem{ShirGecco09}
Shir, O.M., Roslund, J., Rabitz, H.:
\newblock {Evolutionary Multi-Objective Quantum Control Experiments with the
  Covariance Matrix Adaptation}.
\newblock In: Proceedings of the Genetic and Evolutionary Computation
  Conference, GECCO-2009, New York, NY, USA, ACM Press (2009)  659--666

\bibitem{Baeck-book}
B{\"a}ck, T.:
\newblock {E}volutionary {A}lgorithms in {T}heory and {P}ractice.
\newblock Oxford University Press, New York, NY, USA (1996)

\bibitem{HansenDR2PPSN08}
Ros, R., Hansen, N.:
\newblock {A} {S}imple {M}odification in {CMA}-{ES} {A}chieving {L}inear {T}ime
  and {S}pace {C}omplexity.
\newblock In: Parallel Problem Solving from Nature - PPSN X. Volume 5199 of
  Lecture Notes in Computer Science., Springer (2008)  296--305

\bibitem{Ostermeier94}
Ostermeier, A., Gawelczyk, A., Hansen, N.:
\newblock {A Derandomized Approach to Self Adaptation of Evolution Strategies}.
\newblock Evolutionary Computation \textbf{2}(4) (1994)  369--380

\bibitem{Rudolph92}
Rudolph, G.:
\newblock {On Correlated Mutations in Evolution Strategies}.
\newblock In: Parallel Problem Solving from Nature - PPSN II, Amsterdam,
  Elsevier (1992)  105--114

\bibitem{regtutorial}
Neumaier, A.:
\newblock Solving ill-conditioned and singular linear systems: A tutorial on
  regularization.
\newblock SIAM Rev. \textbf{40}(3) (1998)  636--666

\bibitem{NumRecipes}
Press, W.H., Teukolsky, S., Vetterling, W., Flannery, B.:
\newblock Numerical Recipes.
\newblock Cambridge University Press, New York (1992)

\bibitem{Suganthan}
Suganthan, P.N., Hansen, N., Liang, J.J., Deb, K., Chen, Y.P., Auger, A.,
  Tiwari, S.:
\newblock {Problem Definitions and Evaluation Criteria for the CEC 2005 Special
  Session on Real-Parameter Optimization}.
\newblock Technical report, Nanyang Technological University, Singapore (2005)

\bibitem{dusi01}
Dudovich, N., Dayan, B., Gallagher~Faeder, S.M., Silberberg, Y.:
\newblock {Transform-Limited Pulses Are Not Optimal for Resonant Multiphoton
  Transitions}.
\newblock Phys. Rev. Lett. \textbf{86}(1) (2001)  47--50

\bibitem{Roslund06}
Roslund, J., Roth, M., Rabitz, H.:
\newblock {Laboratory Observation of Quantum Control Level Sets}.
\newblock Phys. Rev. A \textbf{74}(4) (2006)  043414

\bibitem{Movie_focal-shg}
Shir, O.M., Roslund, J.:
\newblock {Online Supplementary Material: FOCAL Operating on Total-SHG
  Laboratory Experiments}.
\newblock \url{http://www.princeton.edu/~oshir/focal_shg.zip} (2010)

\end{thebibliography}
